\documentclass[preprints,submit,accept,pdftex,moreauthors]{Definitions/mdpi} 
\firstpage{1} 
\pubvolume{1}
\issuenum{1}
\articlenumber{0}
\pubyear{2025}
\copyrightyear{2025}
\datereceived{ } 
\daterevised{ } 
\dateaccepted{ } 
\datepublished{ } 
\hreflink{https://doi.org/} 

	\usepackage[normalem]{ulem}
        \sethlcolor{lightgray} 
	\definecolor{flame}{rgb}{0.89, 0.35, 0.13}
    
        \usepackage[most]{tcolorbox}
        \newtcolorbox{highlighted}{colback=lightgray,coltext=black,breakable} 
	
	\usepackage{framed}

	\Title{Can Artificial Intelligence Write Like Borges? An Evaluation Protocol for Spanish Microfiction}
	
	\TitleCitation{Can Artificial Intelligence Write Like Borges? An Evaluation Protocol for Spanish Microfiction}
	
	\Author{Gerardo Aleman Manzanarez$^{1}$\orcidA{}, Nora de la Cruz Arana$^{1}$ , Jorge Garcia Flores$^{2}$ , Yobany Garcia Medina$^{1}$, Raul Monroy$^{1}$*\orcidA{}, Nathalie Pernelle$^{2}$}
	
	\AuthorNames{Gerardo Aleman Manzanarez, Jorge Garcia Flores , Nora de la Cruz Arana, Yobany Garcia Medina, Raul Monroy and Nathalie Pernelle}
	
	\isAPAStyle{%
		\AuthorCitation{Aleman, G., de la Cruz, N., Garcia, J., Garcia, Y.,  Monroy, R., \& Pernelle, N.}
	}{%
		\isChicagoStyle{%
			\AuthorCitation{Aleman Manzanarez, Gerardo, de la Cruz Arana, Nora, Garcia Flores, Jorge, Garcia Medina, Yobany, Monroy, Raul, and Pernelle Nathalie.}
		}{
			\AuthorCitation{Aleman Manzanarez, G.; de la Cruz Arana, N.; Garcia Flores, J.; Garcia Medina, Y.; Monroy, R.; Pernelle, N.}
		}
	}
	
	\address{%
		$^{1}$ \quad Tecnologico de Monterrey, Carr.~Lago de Guadalupe Km.3.5, Col.~Margarita M. de Juarez, Atizapan, Estado de Mexico, Mexico\\
		$^{2}$ \quad Centre National de la Recherche Scientifique - Laboratoire d'Informatique de Paris Nord - Université Sorbonne Paris Nord, 
		99 av. Jean-Baptiste Clément, 93430 Villetaneuse, France; jgflores@lipn.univ-paris13.fr}
	
	\corres{Correspondence: raulm@tec.mx}
	
	\abstract{Automated story writing has been a subject of study for over 60 years. Large language models can generate narratively consistent and linguistically coherent short fiction texts. Despite these advancements, rigorous assessment of such outputs for literary merit—especially concerning aesthetic qualities—has received scant attention. In this paper, we address the challenge of evaluating AI-generated microfictions and argue that this task requires consideration of literary criteria across various aspects of the text, such as  thematic coherence, textual clarity, interpretive depth, and aesthetic quality. To facilitate this, we present GrAImes: an evaluation protocol grounded in literary theory, specifically drawing from a literary perspective, to offer an objective framework for assessing AI-generated microfiction. Furthermore, we report the results of our validation of the evaluation protocol, as answered by both literature experts and literary enthusiasts. This protocol will serve as a foundation for evaluating automatically generated microfictions and assessing their literary value.}
	

\begin{document}
		
		
		
	\section{Introduction}
	Technological progress in Artificial Intelligence (AI) has led to systems capable of advanced reasoning \citep{guo2025deepseek, jaech2024openai, team2024gemini}, multimodal understanding \citep{team2024gemini} and creative writing \citep{porter2024ai}. Additionally, improvements in knowledge distillation \citep{hinton2015distilling} and reductions in inference-time computational costs \citep{team2024gemini, leslie2024frontier} suggest that generative AI systems are becoming more accessible and affordable, with some scientific studies suggest that AI-generated texts may surpass human-written ones in literary quality \citep{clark2018neural, jakesch2023human}. However, the assessment of literary quality has always been a subjective matter.
		
		A review of current methods for automatically generating and evaluating automatically generated fiction \citep{alhussain2021automatic} reveals that literary criteria are seldom considered. In particular, the concept of reception \citep{iser1979act, ingarden1993concretizacion} plays a crucial role. As such study indicates, the reception of literary texts is significantly influenced by readers' experiences and literary expertise. Consequently, an evaluation framework grounded in literary theory is essential for effectively comparing human-authored texts with those produced by generative models.
		
                Moreover, beyond a direct human-AI comparison, it becomes relevant to assess the aesthetic quality of AI-generated fiction through a literary evaluation protocol. Current evaluation methods, which mostly come from the fields of Natural Language Processing and Machine Translation, exhibit significant limitations when applied to literary texts. These methods rely heavily on quantitative metrics such as BLEU, ROUGE, and perplexity, which fail to capture nuanced aspects of literary language, like metaphor, symbolism, and stylistic creativity. These metrics prioritize surface-level similarity over deeper semantic and aesthetic qualities, leading to inadequate assessments of text richness. While AI mechanisms can generate consistent and coherent fiction, little attention has been given to producing texts with literary value. This raises the fundamental question: what makes literature literature?
                
                This paper introduces GrAImes (Grading AI and Human Microfiction Evaluation System)\footnote{ \textit{Grading AI Microfiction Evaluation System} evaluation protocol is named after Joseph E. Grimes (born 1922), an American linguist known for his work in discourse analysis, computational linguistics, and indigenous languages of America. He developed the first automatic story generator in the early 1960s~\citep{ryan2017grimes}, a pioneering system that used Vladimir Propp's analysis of Russian folk stories~\citep{propp2012russian} as a grammar for story generation.} a novel evaluation protocol specifically designed to assess the literary value of microfictions. GrAImes incorporates literary criteria into the assessment process, aiming to evaluate the literary quality of microfictions generated by AI, as well as those created by human authors, with or without AI assistance.
        
                GrAImes reception situation is inspired by the editorial process used to accept or reject stories submitted to the publishing industry~\citep{ginna2017editors}. We chose to work with microfictions (see section \ref{microfiction_def}) as a model of literary narrative, both due to the limitations of language models in generating short narratives and because their brevity facilitates the evaluation process. GrAImes consists of a questionnaire with fifteen questions designed to assess literary originality, the impact of microfiction (MF) on its readers, and, to some extent, its commercial potential.

	The protocol’s development involved an initial validation phase, during which a group of experts (literary scholars holding a PhD and an academic position) assessed human-written microfictions. Subsequently, GrAImes was applied to AI-generated microfictions, with evaluations conducted by a community of reading and literature enthusiasts. In both experiments, it is necessary to take into consideration that human evaluators base their assessments on their imagination and prior knowledge, often assigning higher ratings to stories they find more familiar \citep{peinado2006evaluation}. From a historical creativity perspective \citep{boden2004creative}, their evaluations are shaped by comparisons with other narratives they have encountered throughout their lives. With GrAImes we propose an evaluation method that aims to not only to texts likeness but also to the their aesthetic, technical, editorial and commercial quality and value. 	
		
    Our findings indicate that GrAImes could become a reliable framework for assessing the literary quality of human written and AI-generated microfictions. Results from the first experiment, where literature experts evaluated anonymous, human written microfictions, suggest a correlation between the author's expertise and the microfictions' evaluation, with good to acceptable internal consistency in the evaluator's judgment. This was further corroborated in a second experiment involving reading enthusiasts and experts, whose evaluations slightly favored microfictions generated by ChatGPT-3.5 over those produced by a GPT-2 baseline fine-tuned language model (see section \ref{Monterros_system}). The cumulative results of these experiments position GrAImes as a valuable tool for aiding the validation process of microfictions, demonstrating good to acceptable internal consistency in the evaluators' judgments.
		
    However, it is important to acknowledge the limitations of the present study. The experimental dataset of our experiments is small, and some statistical methods used for validation (ICC, Cronbach's alpha, and Kendall's W) are sensitive to the sample size, which might introduce bias or instability in the results. Furthermore, additional research is needed to test the applicability of GrAImes to other literary genres and other languages, as the present experiments were performed only in Spanish. Despite these limitations, we hope that our work shows the value of introducing concepts, methods and expertise from the literary field, especially in order to challenge recent research results \citep{porter2024ai} suggesting that average readers prefer AI-generated poetry to classical texts.
        
        This paper introduces GrAImes, a novel evaluation protocol specifically designed for assesing the literary value of microfictions. The protocol's development involved an initial validation phase where literary experts assessed human-written microfictions. Subsequently, GrAImes was applied to AI-generated microfictions, with evaluations conducted by both literary experts and literature enthusiasts. The cumulative results of these experiments position GrAImes as a valuable tool for aiding the validation process of microfictions.

        The paper is structured to first provide a foundational understanding of the subject. It begins with a comprehensive definition and illustrative examples of microfictions (Section \ref{microfiction_def}). Following this, the evaluation protocol is detailed (Section \ref{sec:subs_experiment}). This section will distinguish between the evaluation processes: literary experts evaluating human-written microfictions, and literature enthusiasts evaluating AI-generated microfictions. The GrAImes evaluation protocol itself is then formally presented. To contextualize the work within the AI text generation field, the paper includes a review of existing approaches for evaluating automatic text generation (Section \ref{sec:related-work}). It then specifies the chatbots used for generating AI microfictions in the experiments, notably the Monterroso system (Section \ref{Monterros_system}) and ChatGPT. Finally, the paper concludes with a thorough presentation and discussion of the results obtained from the conducted experiments, highlighting the insights gained from applying GrAImes to both human and AI-generated microfictions.

	\section{Microfictions and evaluation methods }
        \subsection{Microfictions} 
		\label{microfiction_def}
		Microfiction is a genre that mimics the narrative process through various significant mechanisms: transtextual, metafictional, parodic, and ironic, to construct its structure at both the syntactic and semantic levels. Is an exceptionally brief story with a highly literary purpose ~\citep{tomassini1996minificcion}, far surpassing the ambition of generating readable and coherent narratives. Additionally, this literary genre challenges narrative norms by intentionally disarticulating the plot, requiring the reader's narrative intelligence to navigate. While its brevity may be justified by its limited word count, this characteristic alone does not determine its textual functioning. Instead, microfiction relies on the literary system as its primary reference, offering a reinterpretation of previously explored fictional concepts. It deliberately disrupts its plot, creating gaps in the narrative framework that the reader must fill to engage with the story. Therefore, microfiction cannot be classified solely based on word count but rather through the strategic use of information that prompts transtextual relationships, enabling the reader to reactivate the text's signifying process~\citep{medina2017microrrelato}.
				
\begin{figure}[h!]
			\includegraphics[width=13cm]{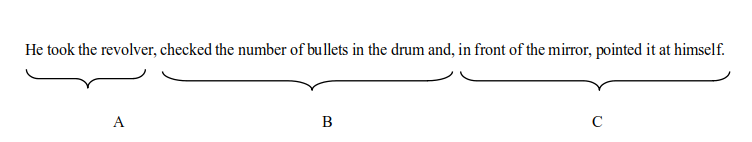}
			\caption{This figure presents an example of a microfiction authored by Yobany Garcia Medina. It is divided into three parts: opening (A), development (B), and closing (C), collectively forming a cohesive narrative unit.}
			\label{fig8:m8}
		\end{figure}

		Microfiction follows a structured sequence: opening, development, and closing. Each action transitions from one state to another — from point A to point C, passing through B— to form a cohesive narrative unit (see Figure~\ref{fig8:m8}).
				
\begin{figure}[h!]
			\includegraphics[width=13cm]{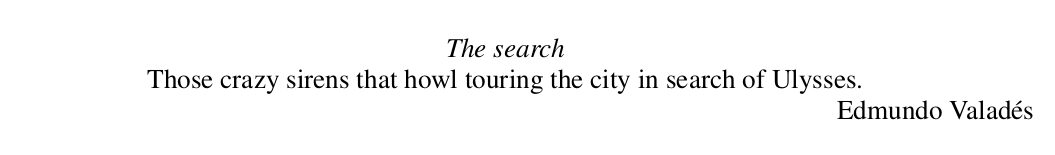}
			\caption{Microfiction example}
			\label{fig11:m11}
		\end{figure}

		Reading microfiction requires the reader not only to interpret its meaning but also to reconstruct its structure. The text prompts the reader to complete the narrative by providing cues that suggest a storyline. As \cite{ricoeur1989funcionnarrativa} suggests, 'fictions re-describe what conventional language has already described.' Therefore, microfictions reinterpret previously explored literary themes. Consequently, while literary forms may differ in nomenclature, they also exhibit structural distinctions and inevitable variations in reception. Refer to Figure~\ref{fig11:m11} for an example of microfiction.
		
		In the realm of microfiction, where syntax is condensed into a dense network of signs, the reader’s role becomes indispensable. While every text requires interpretation, microfiction demands particularly active engagement to decipher its tacitly encoded information. As \cite{barthes1990aventura} define them, these codes—'a system of associative informational fields referring to various spheres of culture'—shape the narrative structure and construct its semantics through a hermeneutical process. This initial dimension shapes the aesthetic experience of the second by introducing interpretative ambiguities. Thus, reading reactivates both explicit textual information and implicit transtextual references.
		
		As we explore in Section~\ref{sec:related-work}, the aesthetic dimension is rarely considered in the design of text evaluation protocols. Therefore, it is essential to consider the aspects of literature in this context. In our study, microfiction is defined as a narrative text limited to 300 words, aligning with the established parameters as articulated by Ana Maria Shua \citep{shua2023como}, characterized by concision, suggestive narrative, and complete story conveyance within a constrained format.

		\subsection{How text generation has been evaluated ?}
		\label{sec:related-work}		
			
			The research on text generation spans a variety of approaches, with each study aiming to advance the field in a unique manner. Table \ref{table:evalmech} presents an overview of different methodologies, highlighting the objectives pursued by various researchers and the evaluation mechanisms used to assess their effectiveness.
			
			\begin{table}[h]	
				\caption{Automatic text generation evaluation methods.} 
				\label{table:evalmech}	
				\tiny
				\begin{center}
					\begin{tabular} {  p{2cm}  p{3.5cm}  p{1.5cm}  p{2.5cm}  p{2cm} }
						\hline
						Author & Goal & Approach & Evaluation method  & Results \\ [0.5ex] 
						\hline
						\citeauthor{fan2018hierarchical}, \citeyear{fan2018hierarchical} & Hierarchical story generation using a fusion model  & Deep Learning & Human evaluation, Perplexity & Story generation with a given prompt.\\
						\hline
						\citeauthor{guan2021long}, \citeyear{guan2021long} & Long text generation & Deep Learning & Perplexity, BLEU, Lexical Repetition, Semantic Repetition, Distinct-4, Context Relatedness, Sentence Order & Generation of long texts using sentence and discourse coherence. \\
						\hline
						\citeauthor{min2021deep}, \citeyear{min2021deep} & Short text generation for an image & Deep Learning & None  & Generation of short texts using an image and encoder-decoder structure. \\
						\hline
						\citeauthor{lo2022gpoet}, \citeyear{lo2022gpoet} & Poem generation using GPT-2 & Deep Learning & Lexical diversity, Subject continuity, BERT-based Embedding Distances, WordNet-based Similarity Metric, Content Classification & Limerick poems with AABBA rhyming scheme.\\
						\hline
						\citeauthor{cavazza2002character}, \citeyear{cavazza2002character} & Character-based interactive story generation & Rule-based  & Quantification of system’s generative potential & Computer entertainment story generation. \\
						\hline
						\citeauthor{gervas2005story}, \citeyear{gervas2005story} & Story that matches a given query & Ontology-based & None & Sketch of a story plot. \\ 
						\hline
						\citeauthor{mori2019toward}, \citeyear{mori2019toward} & Story generation with better story endings & Neural Network-based & Human evaluation & Endings containing positive emotions, supported by sentiment analysis. \\
						\hline
						\citeauthor{rishes2013generating}, \citeyear{rishes2013generating} & Story generation & Symbolic Encoding & Levenshtein Distance, BLEU & Stories and fables.\\
						\hline
						\citeauthor{elson2009tool}, \citeyear{elson2009tool} & Annotation tool for the semantic encoding of texts & Symbolic Encoding & Human evaluation & Short fables.\\
						\hline
						\citeauthor{sutskever2011generating}, \citeyear{sutskever2011generating} & Character-level language modeling & Neural Network-based & Bits Per Character & Text generation with gated RNNs.\\
						\hline 
						\citeauthor{kiddon2016globally}, \citeyear{kiddon2016globally} & To generate an output by dynamically adjusting interpolation among a language model and attention models & Neural Network-based & Human evaluation, BLEU-4, METEOR & Text generation with global coherence.\\
						\hline
						\citeauthor{zhu2015aligning}, \citeyear{zhu2015aligning} & Rich descriptive explanations and alignment between book and movie & Neural Network-based & BLEU, TF-IDF  & Story generation from images and texts.\\
						\hline
						\citeauthor{walker2011perceived},  \citeyear{walker2011perceived}& Generate story dialogues from film characters & Statistical Model & Human evaluation & Dialogues generated based on given film characters.\\
						\hline
						\citeauthor{sharp2016sunspring}, \citeyear{sharp2016sunspring} & Generation of a film script & Neural Network-based & None  & Short film script.\\
						\hline 
						\citeauthor{lukin2017generating}, \citeyear{lukin2017generating} & Sentence planning & Neural Network-based & Levenshtein Distance, BLEU & Parameterized sentence planner.\\
						\hline
                        \citeauthor{porter2024ai},\citeyear{porter2024ai} & Poem generation using ChatGPT & Deep Learning & Human (crowd evaluation with Prolific) & Evaluators preferred ChatGPT poems to those from well known human authors.\\
                        \hline
					\end{tabular}
				\end{center}
			\end{table}

			In the domain of narrative fiction generation, \cite{fan2018hierarchical} focuses on generating structured narratives using a fusion model, ensuring coherence across different hierarchical levels. The evaluation of this approach involves human evaluation and perplexity measures. Similarly, \cite{guan2021long} aims to improve long-text generation by enhancing sentence and discourse coherence, employing deep learning (DL) techniques. The evaluation metrics used include perplexity, BLEU (Bilingual Evaluation Understudy) \citep{papineni2002bleu}, lexical and semantic repetition, distinct-4, context relatedness, and sentence order, which collectively assess the model’s ability to maintain logical flow and coherence.
			
			Shorter-form text generation has also been explored, with \cite{min2021deep} proposing an encoder-decoder structure for generating short texts based on images, optimizing for succinctness and relevance. Notably, this approach lacks an explicit evaluation mechanism. Meanwhile, \cite{lo2022gpoet} utilizes GPT-2 \citep{radford2019language} to generate structured poetry, adhering to specific poetic constraints such as the AABBA rhyming scheme of limerick poems. The effectiveness of this approach is evaluated using lexical diversity, subject continuity, BERT-based \cite{devlin2019bert} embedding distances, WordNet-based similarity metrics, and content classification.
			
			Beyond traditional story generation, some studies focus on interactive and rule-based systems. \cite{cavazza2002character} introduces a character-based interactive storytelling mechanism aimed at enhancing computer entertainment, evaluated through a quantification of the system’s generative potential. \cite{gervas2005story}, employing an ontology-based system to generate plots matching specific user queries, does not specify an evaluation mechanism.
			
			Further refining narrative outcomes, \cite{mori2019toward} investigates the impact of sentiment-driven story endings, leveraging neural networks to generate positive emotional resolutions. The evaluation relies on human judgment to assess the effectiveness of emotional storytelling. Similarly, \cite{rishes2013generating} develops a symbolic encoding method for automated story and fable creation, evaluated using Levenshtein distance and BLEU scores to measure textual similarity and fluency.
			
			Annotation tools also play a role in text generation research, as exemplified by \cite{elson2009tool}, which introduces a semantic encoding system designed for textual annotation. This study is evaluated using human assessment to verify the quality of semantic encodings. In a different vein, \cite{sutskever2011generating} explores character-level language modeling with gated recurrent neural networks (RNNs) to improve text synthesis at a granular level, employing bits per character as the primary evaluation metric.
			
			Efforts in coherence and interpolation-based generation are represented by \cite{kiddon2016globally}, who propose a model that dynamically adjusts interpolation between a language model and attention mechanisms to maintain global coherence. The evaluation relies on human judgment as well as BLEU-4 and METEOR (Metric for Evaluation of Translation with Explicit Ordering) scores to assess linguistic accuracy and coherence. Additionally, \cite{zhu2015aligning} investigates alignment techniques for rich descriptive explanations, aiming to generate textual content that effectively bridges books and their movie adaptations. This approach is evaluated using BLEU and TF-IDF (Term Frequency–Inverse Document Frequency) similarity measures.
			
			Some researchers focus on dialogue-based storytelling. \cite{walker2011perceived} develops a statistical model to generate film dialogues based on character archetypes, ensuring that generated dialogues maintain consistency with established personas. The evaluation relies on human assessments of the generated dialogues. Expanding upon this, \cite{sharp2016sunspring} explores automated scriptwriting using neural networks, resulting in the creation of a short film script; however, no explicit evaluation mechanism is reported. Other approaches like \cite{lukin2017generating} delves into sentence planning techniques, employing neural network-based methodologies to refine parameterized sentence structure generation. The evaluation uses Levenshtein distance and BLEU scores to measure textual structure and fluency.

                        Lastly, a critical stand is needeed concerning recent research results from \citep{porter2024ai} suggesting that average readers prefer AI-generated poetry to classical texts. While the statistical rigor and scale of this study are noteworthy, its interpretive claims are weakened by a fundamental oversight: the absence of literary concepts, particularly the theory of reception \citep{ingarden1993concretizacion, iser1979act}. By relying exclusively on a crowd-sourced evaluation platform populated by self-reported non-experts (90.4\% of whom read poetry infrequently and two-thirds of whom were unfamiliar with the assigned poets) the authors equate statistical significance with literary insight. Reception theory emphasizes that meaning is co-produced by readers whose interpretive frameworks are shaped by their literary competence, cultural background, and historical awareness \cite{ricoeur1989funcionnarrativa, barthes1990aventura}. The study’s conclusion that AI-generated poems are preferred over canonical works thus rests on a narrow understanding of preference that neglects the interpretive depth and aesthetic endurance that define literary reception. Without accounting for the respondents’ literary literacy, the finding that AI poems are judged superior risks privileging superficial readability over poetic complexity, potentially reframing the appreciation of literature in terms of immediate appeal rather than interpretive depth and enduring value.
			
			Overall, these diverse research efforts illustrate the breadth of text generation methodologies, encompassing deep learning, rule-based systems, and symbolic encoding, each targeting unique challenges in narrative coherence, stylistic constraints, and interactivity. The evaluation mechanisms vary widely, with some studies relying on automated metrics such as BLEU and perplexity, while others emphasize human evaluation to assess narrative quality and coherence.
			
			\subsection{Literary text evaluation methods}
			\label{sec:Lit_txt_eval_methds}
			
			The evaluation of literary text generation remains an open challenge, rooted in longstanding literary traditions and formalist approaches \citep{propp2012russian, genetteFigures1976, bal2009narratology}. Despite the growing interest in computational creativity, there is no clear consensus on how to effectively assess creative text generation or measure the contribution of different stages in the process. These stages range from knowledge-based planning \cite{gardent2017statistical} to structuring the temporal flow of events \citep{ Dorr07summarization-inspiredtemporal-relation} and producing linguistic realizations. Among these sub-tasks, evaluation is arguably the least developed and requires further research efforts \citep{zhu2012evaluation}. The widespread adoption of commercial Large Language Model based chatbots has resulted in increased scientific efforts into the evaluation of AI-generated text reception, with a particular emphasis on human evaluation methodologies. Notable contributions in this area include the works of \citeauthor{porter2024ai} \cite{porter2024ai}, \citeauthor{koziev2025automatedevaluationmeterrhyme} \cite{koziev2025automatedevaluationmeterrhyme}, and \citeauthor{franceschelli2025thinkingoutsidegraybox} \cite{franceschelli2025thinkingoutsidegraybox}.
											
			In summary, the evaluation of literary text generation remains a complex and evolving challenge. While human evaluation provides the most reliable assessments, untrained and machine-learned metrics offer scalable alternatives with varying degrees of effectiveness. Future research must focus on refining these methodologies to better capture the nuances of creative and narrative text generation.
			\section{Materials and Methods}

            \subsection{Evaluation Protocol}
			It is important to note that the definition of literarity is ideological and sociohistorical, hence not fixed in time but embedded in a cultural context~\citep{ludmer2015clases}. In educational settings applying the communicative approach to language, drawn from linguistic pragmatics~\citep{austin1962austin}, literature is characterized as a form of communication distinguished by four elementary features.
			
			The first feature is verisimilitude; a literary text is grounded in everyday reality but instead of replicating it, it represents it. Consequently, it does not rely on external references but rather creates them, demanding readers to engage in a cooperative pact accepting the proposed universe as plausible. The second distinguishing aspect of literature is its codification. It is a message where every component is chosen to convey meaning, with each perceived as intentional and linked to the total meaning of the work. Thus, a literary text can not be summarized, translated, or paraphrased without significant loss of its essence~\citep{bertochi1995aproximacion}.
			
			Derived from this, the third feature is the deliberate breaking of rules and conventions of everyday language, and even strict grammar, in favor of aesthetic effect or meaning. This creates a tension between literature and language, emphasizing how something is said over what is said. Lastly, the deferred character of literarity~\citep{huaman2003educacion} refers to the fact that sender and receiver of a work rarely share context, influencing reception but not decisively for understanding the text. Autonomy largely depends on the integrity and cohesion of the diegetic world.
			
			Yet, one question remains: do these elements are enough to consider a text as literary? Functionally, perhaps, but literary communication holds a significant ideological component dependent on sociohistorical context. In cultural studies, a distinction between ``literary” and ``consumer“ fiction is made based on one variable: prestige~\citep{de2002telenovela}. Traditionally, canonical literature was determined by academia or critics, while the publishing industry, since the 19th century, played a pivotal role in validation~\citep{thompson2013merchants}. Each participant uses different parameters and perspectives; however, the editor's role, mediating between author, reader, and time~\citep{calasso2014marca}, is arguably the most operational and inclusive. Editors assess content clarity, technical value based on genre, and relevance, which can be thematic, formal, or commercial~\citep{ginna2017editors}. Hence, these three parameters --- clarity, technical value, and relevance --- applied to microfiction evaluation, gauge their potential for publication and integration into the contemporary literary landscape.
			
			The initial assessment for the publication of an unsolicited manuscript by a publisher typically involves an evaluation process known as opinion. This usually entails a report prepared by a specialized reader, focusing on the content's technical value, commercial potential, and potential marketing strategies. To address these aspects systematically, we propose an evaluation instrument for microfictions, consisting of questions that can be answered by both specialized and non-specialized readers.

			The evaluation framework outlined in Table \ref{table:GrAImes_questions} presents the GrAImes evaluation protocol, which includes 15 items specifically tailored to assess Spanish microfictions, with broader applicability to narrative productions across genres and languages (for which further research is needed).

                        The protocol is organized into three distinct dimensions, each addressing specific criteria for systematic analysis of the texts assigned to evaluators. The first dimension, labeled "Story Overview and text complexity," evaluates literary quality through an assessment of thematic coherence, textual clarity, interpretive depth, and aesthetic merit, incorporating both quantitative metrics (e.g., scoring scales) and qualitative judgments (e.g., textual commentary) to appraise literary value. The second dimension, "Technical Evaluation," focuses on technical aspects such as linguistic proficiency, narrative plausibility, stylistic execution, genre-specific conventions, and the effective use of language to convey meaning. The final dimension, "Editorial / Commercial quality," examines the commercial potential and editorial suitability of the microfictions, assessing factors such as audience appeal, market relevance, and feasibility for publication or dissemination. This tripartite structure ensures a comprehensive, multidimensional evaluation of both artistic and practical qualities inherent to the microfiction genre.
			
			\begin{table}[h]
				\scriptsize
				\caption{List of questions in the evaluation protocol provided to the evaluators tasked with assessing the literary, linguistic, and editorial quality of microfiction pieces. OA = Open Answer, Likert = Likert's scale from 1 to 5.}
				\label{table:GrAImes_questions}
				\begin{tabularx}{\textwidth}{p{0.05cm} p{5cm} p{.5cm} X}
					\hline
					\multicolumn{4}{c}{\textbf{GrAImes Evaluation Protocol Questions}} \\
					\hline
					\textbf{\#} & \textbf{Question} & \textbf{Answer} & \textbf{Description} \\
					\hline
					\multicolumn{4}{c}{\textbf{Story Overview and text complexity}} \\
					\hline
					1 & What happens in the story? & OA & Evaluates how clearly the generated microfiction is understood by the reader. \\
					\hline
					2 & What is the theme? & OA & Assesses whether the text has a recognizable structure and can be associated with a specific theme. \\
					\hline
					3 & Does it propose other interpretations, in addition to the literal one? & Likert & Evaluates the literary depth of the microfiction. A text with multiple interpretations demonstrates greater literary complexity. \\
					\hline
					4 & If the above question was affirmative, Which interpretation is it? & OA & Explores whether the microfiction contains deeper literary elements such as metaphor, symbolism, or allusion. \\
					\hline
					\multicolumn{4}{c}{\textbf{Technical Assessment}} \\
					\hline
					5 & Is the story credible? & Likert & Measures how realistic and distinguishable the characters and events are within the microfiction. \\
					\hline
					6 & Does the text require your participation or cooperation to complete its form and meaning? & Likert & Assesses the complexity of the microfiction by determining the extent to which it involves the reader in constructing meaning. \\
					\hline
					7 & Does it propose a new perspective on reality? & Likert & Evaluates whether the microfiction immerses the reader in an alternate reality different from their own. \\
					\hline
					8 & Does it propose a new vision of the genre it uses? & Likert & Determines whether the microfiction offers a fresh approach to its literary genre. \\
					\hline
					9 & Does it give an original way of using the language? & Likert & Measures the creativity and uniqueness of the language used in the microfiction. \\
					\hline
					\multicolumn{4}{c}{\textbf{Editorial / Commercial Quality}} \\
					\hline
					10 & Does it remind you of another text or book you have read? & Likert & Assesses the relevance of the text and its similarities to existing works in the literary market. \\
					\hline
					11 & Would you like to read more texts like this? & Likert & Measures the appeal of the microfiction and its potential marketability. \\
					\hline
					12 & Would you recommend it? & Likert & Indicates whether the microfiction has an audience and whether readers might seek out more works by the author. \\
					\hline
					13 & Would you give it as a present? & Likert & Evaluates whether the microfiction holds enough literary or commercial value for readers to gift it to others. \\
					\hline
					14 & If the last answer was yes, to whom would you give it as a present? & OA & Identifies the type of reader the evaluator believes would appreciate the microfiction. \\
					\hline
					15 & Can you think of a specific publisher that you think would publish a text like this? & OA & Assesses the commercial viability of the microfiction by determining if respondents associate it with a specific publishing market. \\
					\hline
				\end{tabularx}
			\end{table}

            		\begin{figure}[]
                        \caption{Microfiction evaluation process}
			\label{fig10:m10}
                                        \vspace{.2 cm}
		\includegraphics[width=10cm]{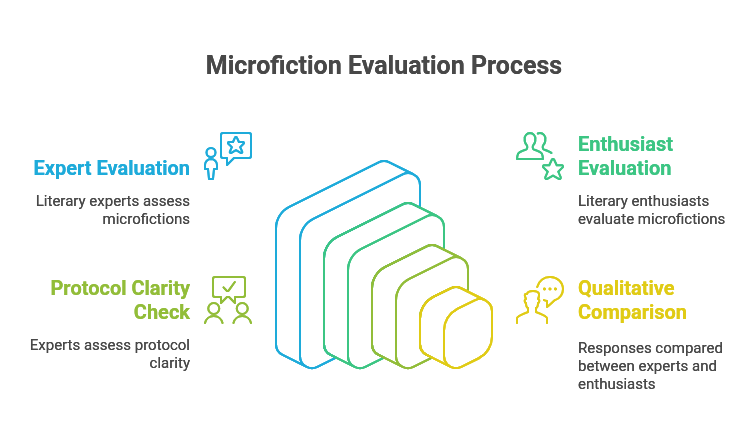}
					\end{figure}
			\subsection{Experiments}\label{sec:subs_experiment}
To test GrAImes' validity, we conducted two experiments. In the first, GrAImes was applied to stories written by humans and evaluated by university experts in literary studies. In the second, we applied GrAImes to stories generated by language models, evaluated by a community of reading and literature enthusiasts. 
                        \subsubsection{Evaluators}
                        
            We gathered two groups of evaluators: the Experts and the Enthusiasts. The selection criteria for the Expert group consisted of five literary scholars, each holding a PhD in Spanish or Latin American literature and occupying a permanent academic position at a public university. The participants were affiliated with institutions in Mexico, France, and the United States. All experts teach literature at the graduate level and are fluent in Spanish. On the other hand, the literature enthusiast group comprised 16 evaluators recruited from a reading club of literature enthusiasts who actively shared their opinions through a YouTube channel and a Telegram group. This group is led by a published Mexican writer and \textit{booktuber}.
        \subsubsection{Datasets}
        The microfiction evaluation dataset for the Experts group consists of six microfictions written in Spanish by human authors (see Table~\ref{table:mfs_authors_experiment1}). We presented these microfictions to the experts along with the fifteen questions from the GrAImes evaluation protocol. The six
			microfictions included two written by an expert and well-known author with published books (MF 1
			and 2), two by a medium experience author who has been published in magazines and anthologies (MF 3 and
			6), and two by an emerging writer (MF 4 and 5).
			Two extra questions were applied to the literary experts evaluating human written microfiction, these questions are: Is this microfiction evaluation protocol clear enough for you? (Yes or No answer); Do you think that this protocol can be used to evaluate the literary value of microfiction?. \label{two_extra_questions}
\begin{table}[h]
        \begin{center}
        \scriptsize
        \caption{{Authors of human written microfictions evaluated by the Expert group}}
        \label{table:mfs_authors_experiment1}
            \begin{tabular}{ |p{1cm} |p{5 cm} |p{2 cm}|}
            \hline
                  \textbf{Author} & \textbf{Experience} & \textbf{Microfictions}\\
                 \hline
                  Expert & Well known with books published & MF1, MF2\\
                 \hline
                  Medium  & Published in magazines and anthologies & MF3, MF6\\
                 \hline
                  Emerging & Little experience, starting author & MF4, MF5\\
                 \hline
            \end{tabular}
        \end{center}
\end{table}

The microfiction evaluation dataset for the Enthusiasts group was generated by two distinct generative AIs: a state-of-the-art large language model, ChatGPT, and an in-house baseline model (Monterroso), specifically trained on a hand-crafted dataset of Spanish microfictions using the GPT-2 architecture. This hand-curated dataset of Spanish microfictions fine-tunes the Monterroso baseline, with the expectation that it would be better attuned to the structural, thematic, and linguistic elements prevalent in this literary form.

The Enthusiasts dataset consists of six AI generated microfictions in Spanish, three microfictions generated by ChatGPT-3.5 and three generated by Monterroso. The Enthusiast group was composed of 16 literary enthousiasts, plus the group leader and \textit{booktuber}. Each evaluator was assigned six different microfictions, in order to answer GrAImes questionnary.
\subsubsection{Microfiction generation systems}
        \paragraph{\textbf{Monterroso}}\label{Monterros_system}			
			Most story generation systems focus on developing a structured framework of narrative elements—such as narrator, character, setting, and time—to enhance story coherence and verisimilitude~\citep{bremond1980logic}. However, they often overlook what \cite{shklovsky1917art} terms “singularization” and what post-structuralist theorists describe as “literariness.”
            The Monterroso baseline system consists of fine-tuning an existing language model with microfictions, in this case, we utilize GPT-2 \citep{radford2019language} as the base model, employing the Deep Learning transformer architecture
            \citep{vaswani2017attention} for training and validation. Monterroso is available in in Spanish and English. With these pretrained models and a hand made corpora of microfictions, Monterroso can produce literary-specific content.
            
            Using the resulting Monterroso model, we input a prompt word, which serves as the title. Additionally, we specify the desired length of the microfiction, with a maximum of 300 words. Monterroso GPT-2 baseline microfictions were used in our experiments. To develop the Monterroso model in Spanish, we leveraged a corpus of 1,33 Spanish microfictions, 1,222 for training and 155 to validate, 1.4 MB size and 411,287 tokens to generate the language model, alongside a publicly available GPT-2 language model specifically tailored for Spanish \citep{GPT2spanish}. 
		\paragraph{\textbf{ChatGPT-3.5}}			
                ChatGPT 3-5~\citep{brown2020gpt3} was used to generate 300 words Spanish microfictions with the same prompts given to the Monterroso baseline system. 
\subsubsection{Statistical measures}
			Regarding the reliability of the evaluation protocol questions, this was assessed using the Intraclass Correlation Coefficient (ICC), which measures the degree of consistency or agreement among responses. Higher ICC values indicate strong reliability, while lower or negative values suggest inconsistencies in response patterns. Additionally, the average scores (AV) provide insight into the perceived difficulty or clarity of each question. The internal consistency of the responses by microfiction was evaluated using Cronbach’s Alpha, with the values categorized into standard reliability thresholds. Given the sensitivity of both Intraclass Correlation Coefficient and Cronbach's Alpha to sample size, we used Kendall's W to evaluate the concordance coefficient. Kendall's W is more appropriate for smaller sample sizes, ensuring a more robust assessment of inter-rater agreement in our study.

        Intraclass Correlation Coefficient:
        \begin{linenomath}
            \begin{equation}
                    ICC = \frac{\sigma_b^2}{\sigma_b^2 + \sigma_w^2}
            \end{equation}
        \end{linenomath}
        where: $\sigma_b$ is the variance between subjects and $\sigma_w$ is the variance within subjects. 

        Cronbach's Alpha:
        \begin{linenomath}
            \begin{equation}
                \alpha = \frac{p}{p-1}\left( 1 - \frac{\sum \sigma_{ii}}{\sum \sigma_{ii} + 2\sum_{i<j} \sigma_{ij}} \right)
            \end{equation}
        \end{linenomath}
        where: $p$ is the number of items in the scale, $\sigma_{ii}$ is the variance of item $i$, $\sigma_{ij}$ is the covariance between items $i$ and $j$.

        Kendall W: 
    \begin{linenomath}
        \begin{equation}
            W = \frac{12S}{m^2(n^3-n)}
        \end{equation}
    \end{linenomath}
            where: n is the number of objects, m is the number of judges and S is the sum of ranks squared deviations.
        In this study, we aimed to evaluate the literary quality of Spanish-language microfiction, focusing on both human and AI-generated texts. The purpose was to assess the effectiveness of a standardized evaluation protocol (GrAImes) in capturing literary value across different types of microfiction, including those written by established and emerging authors as well as those generated by AI models. To achieve this, we selected two distinct generative AIs—ChatGPT and Monterroso, a baseline model trained specifically on a curated dataset of Spanish microfictions. This decision was made to compare the quality of texts generated by state-of-the-art language models and models fine-tuned to the specific structural, thematic, and linguistic elements characteristic of Spanish microfiction. We aimed to explore whether such AI-generated texts could hold up to human evaluations in terms of literary quality and whether an evaluation protocol designed for human-written microfiction could be effectively applied to AI-generated works.

        \subsection{Limitations}
            \label{limitations}
                
            Since both the Intraclass Correlation Coefficient and Cronbach’s Alpha are sensitive to sample size, their results may introduce bias or instability in our conclusions. To address this, we incorporated Kendall’s W, a measure less affected by sample size, to assess inter-annotator agreement. Although annotators’ aesthetic judgments may vary due to individual differences in reception and its corresponding biases, the agreement observed with ICC analysis is still verified, where texts authored by more experienced writers receive higher scores.

            Further experiments are required to assess the applicability of the evaluation protocol outside the microfiction literary genre and to microfictions in other languages. Involving literary experts in each target language may be essential to mitigate cross-cultural validity issues, particularly when translating the questions from Spanish into other languages.

        \subsection{Repeatability}
        \label{sec:repeatability}
            All elements to reproduce the experiment can be found in \url{https://github.com/Manzanarez/GrAImes}.
			\section{Results}			
			\subsection{GrAImes evaluation of human written microfictions by the Expert group}
	        GrAlmes was evaluated by literary experts, all Spanish speakers, with one non-native speaker among them. We selected six microfictions written in Spanish and provided them to the experts along with the fifteen questions from our evaluation protocol. The six microfictions included two written by a well-known author with published books (MF 1 and 2), two by an author who has been published in magazines and anthologies (MF 3 and 6), and two by an emerging writer (MF 4 and 5). 

            \begin{figure}
                \centering
                \caption{Expert group evaluation averages
                }
                \vspace{.2 cm}
                \label{fig:lit_exps_beeswarm_hw_mfs}
                \includegraphics[width=0.7\linewidth]{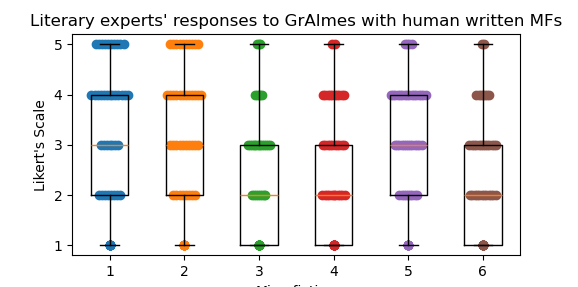}
            \end{figure}

            \begin{figure}
                   \caption{{Expert group evaluation of human written microfictions using GrAImes (averages by section)}}
                   \vspace{.2 cm}
                   \centering
                   \label{fig:lit_exps_bargrph_ai_mfs}\includegraphics[width=0.4\linewidth]{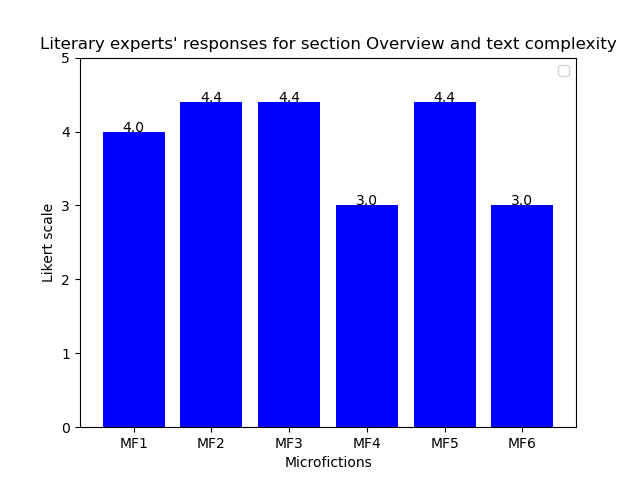}
                \includegraphics[width=0.4\linewidth]{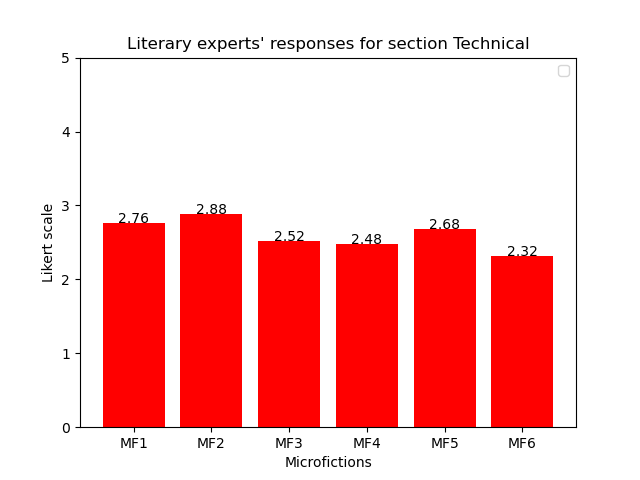}
                \includegraphics[width=0.4\linewidth]{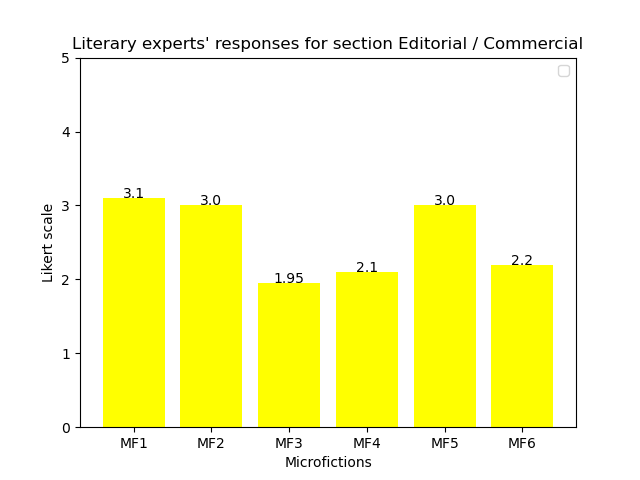}
                \includegraphics[width=0.4\linewidth]{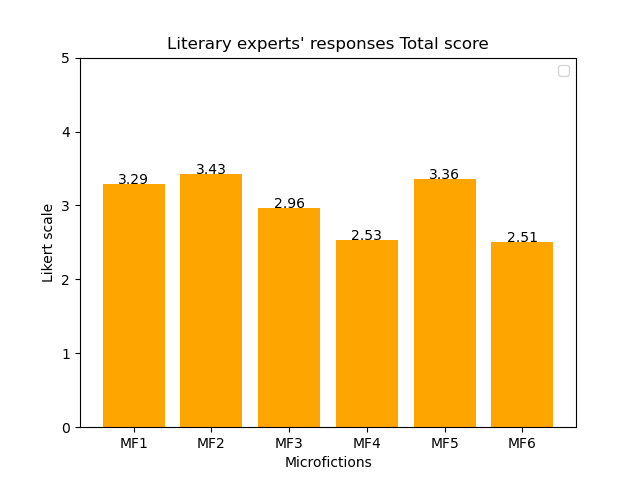}
            \end{figure}

            From the responses obtained and displayed in Tables \ref{table:exp1humanw} and \ref{table:lit_exps_quest_SD_humanw}, we conclude that literary experts rated the microfictions (1 and 2) authored by an expert writer more favorably. However, the responses show a high standard deviation, indicating that while the evaluations were generally positive, there was significant variation among the experts. Additionally, the lowest-ranked microfictions (4 and 6), which have a lower response average, also exhibit a lower standard deviation, suggesting greater agreement among the judges. These texts were written by an emerging author (MF 4) and by a medium experience author (MF 6).
			From the responses obtained and displayed in Tables , we conclude that literary experts rated the microfictions (1 and 2) written by an expert author higher. However, the responses show a high standard deviation, indicating that while the evaluations were generally positive, there was significant variation among the experts. Additionally, the lowest-ranked microfictions (3 and 6), which have a lower response average, also exhibit a lower standard deviation, suggesting greater agreement among the judges. These texts were written by an emerging author published in literary magazines or by small-scale editorial presses with limited book printings.
			\begin{table}[h]
				\scriptsize		
				\caption{Responses, AV and SD, of Literary Experts  to Human-Written Microfictions (MF1-MF6), Measured on a Likert Scale and Grouped by GrAImes Questions Sections, with a Total Average Responses Final Column.}
                
				\label{table:exp1humanw}
				\begin{tabular}{
						m{20em} m{.5em}m{.5em}
						m{.5em} m{.5em} m{.5em} m{.5em} 
						m{.5em} m{.5em} m{.5em} m{.5em}
						m{.5em} m{.5em} c c
					} 
					\hline
					\multicolumn{15}{c}{\textbf{Literary experts responses to human written microfictions}}\\
					\hline
					& \multicolumn{2}{c}{\textbf{MF 1}} & \multicolumn{2}{c}{\textbf{MF 2}}& \multicolumn{2}{c}{\textbf{MF 3}} & \multicolumn{2}{c}{\textbf{MF 4}}& \multicolumn{2}{c}{\textbf{MF 5}} & \multicolumn{2}{c}{\textbf{MF 6}} & \multicolumn{2}{c}{\textbf{Average}}\\
					\hline
					Question & AV & SD & AV & SD & AV & SD  & AV & SD & AV & SD& AV & SD & AV & SD \\
					\hline
					\multicolumn{15}{c}{ \textbf{Story Overview and text complexity}}\\
					\hline
					3.-Does it propose other interpretations, in addition to the literal one? & 4 & 1 & 4.4 & 0.9 & 2.2 & 0.8 & 2.4 & 1.4 & 4.4 & 0.5 & 3.4 & 1.6 & 3.5 &1\\ 
					\hline
					\multicolumn{15}{c}{\textbf{Technical}} \\
					\hline
					5.-Is the story credible?& 2.2 & 1.8 & 3.2 & 1.8 & 4 & 0.5 & 4.4 & 1.7 & 3.4 & 1.8 & 2 & 0.9 & 3.2 &1.4\\
					\hline
					6.-Does the text require your participation or cooperation to complete its form and meaning? & 4.4 &0.9 & 3.6 & 1.3 & 2.6 & 1.5 & 3 & 0.9 & 3.2 & 0.9 & 3.8 & 1.1 & 3.4 &1.1\\
					\hline
					7.-Does it propose a new vision of reality? & 2.4 & 1.1 & 2.6 & 1.5 & 1.2 & 0.9 & 1.8 & 1.1 &3 & 1 & 2.2 & 1.2 & 2.2 &1.1\\
					\hline
					8.-Does it propose a new vision of the genre it uses? & 2 &1.2 & 2.4 &1.5 & 1.4 & 0.5 & 1.2 & 0.4 & 2.2 & 1.1 & 1.6 & 0.9& 1.8 & 0.9\\  
					\hline
					9.-Does it propose a new vision of the language itself? & 2.8 &1.8 & 2.6 & 2.2 & 2.2 &0.9 & 2.8 &1.3 & 1.4 &0.5 & 2 &1.4 & 2.3 &1.4\\  
					\hline
					\multicolumn{15}{c}{\textbf{Editorial / commercial}}  \\
					\hline
					10.-Does it remind you of another text or book you have read? & 4.4 & 0.5 & 4 & 1.1 & 3.2 & 0.4 & 2.8 & 0.8 & 3.6 & 0.4 & 3.4 & 0.8& 3.6 & 0.7\\  
					\hline
					11.-Would you like to read more texts like this? & 3 & 1 & 3 & 0.7 & 1.4 & 0.9 & 2 & 0.8 & 3 & 0.4 & 2 & 0.8 & 2.4 &0.8\\ 
					\hline 
					12.-Would you recommend it? & 2.8 & 1.6 & 3 & 1.2 & 1.2 & 0.9 & 2 & 0.8 & 2.8 & 1.1 & 1.6 & 1 & 2.2 & 1.1\\
					\hline
					13.-Would you give it as a present?  & 2.2 & 1.6 & 2.4 & 1.3 & 1 & 0.9 & 2.2 & 1.2 & 2.4 & 1.1 & 1.8 & 0.8 & 2 & 1.2\\ 
					\hline		
				\end{tabular}
			\end{table}
			
\begin{figure}[]
				\caption{ICC and Cronbach's Alpha line charts of the Expert group evaluation of human written microfictions.}
				\vspace{.2 cm}
				\label{fig:ICC_lit_exp_mf_humw}			\includegraphics[width=.4\linewidth]{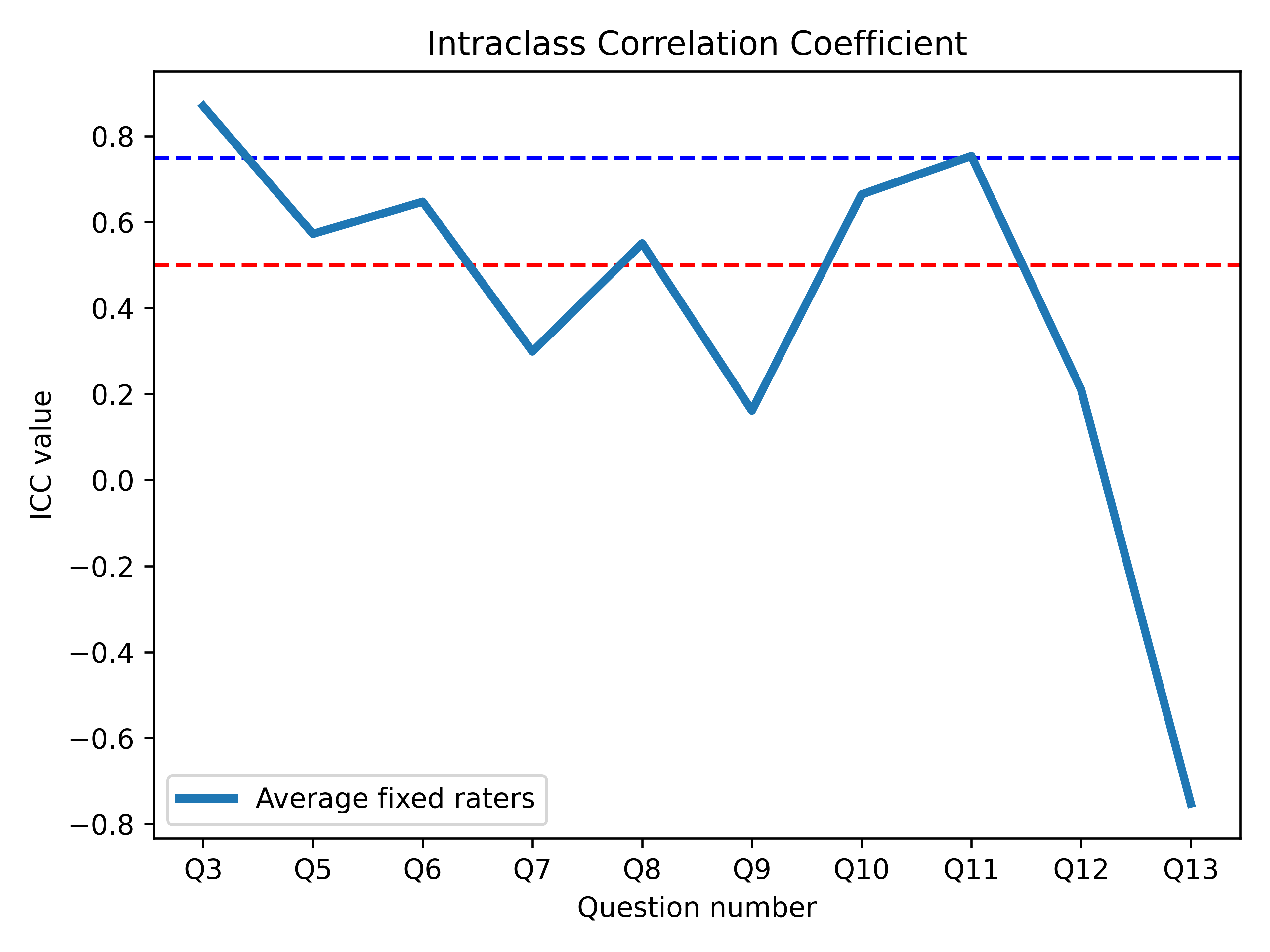}
				\includegraphics[width=.4\linewidth]{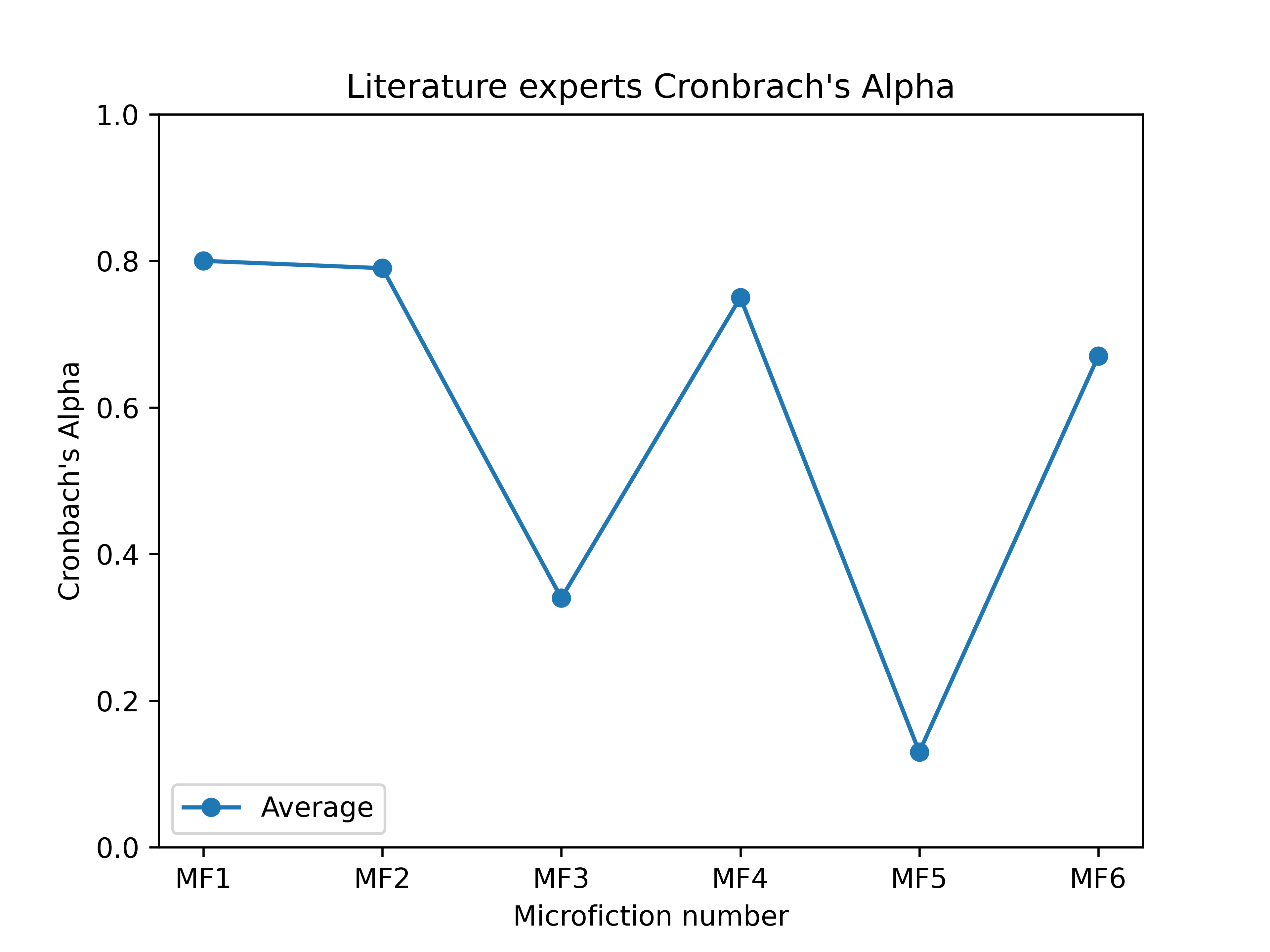}
			\end{figure}

			\begin{table}{}
				\scriptsize	
				\caption{Responses of Literary Experts to GrAImes Questions Evaluating Human Written Microfictions, Organized by Ascending Order of Standard Deviation.}
				\label{table:lit_exps_quest_SD_humanw}
				\begin{tabular}{
						m{40 em} c c
					} 
					\hline
					\multicolumn{3}{c}{\textbf{Literary experts' responses to microfictions written by humans, ordered by SD}}\\
					\hline
					\textbf{Question} 
					& \textbf{AV} & \textbf{SD} \\
					\hline			
					10.-Does it remind you of another text or book you have read? 
					& 3.6 &0.7\\  
					\hline
					11.-Would you like to read more texts like this? 
					& 2.4 &0.8\\ 
					\hline
					8.-Does it propose a new vision of the genre it uses? 
					& 1.8 &0.9\\
					\hline
					3.-Does it propose other interpretations, in addition to the literal one? 
					& 3.5 &1\\
					\hline
					6.-Does the text require your participation or cooperation to complete its form and meaning? 
					&3.4 &1.1\\
					\hline
					7.-Does it propose a new vision of reality?  
					& 2.2 &1.1\\
					\hline 
					12.-Would you recommend it? 
					& 2.2 &1.1\\
					\hline
					13.-Would you give it as a present?  
					&2 &1.2\\
					\hline  
					5.-Is the story credible?
					& 3.2 & 1.4\\
					\hline
					9.-Does it propose a new vision of the language itself? 
					& 2.3 & 1.4\\  
					\hline				
				\end{tabular}
			\end{table}
                        The results suggest a direct correlation between the authors’ expertise and the internal consistency of the texts. The microfictions authored by an expert (MF 1 and MF 2) exhibited the highest Alpha values, 0.80 and 0.79, respectively, indicating good to acceptable internal consistency (see Table (see table \ref{table:ICCAlpha_hwmf} and Figure \ref{fig:ICC_lit_exp_mf_humw}). This suggests that the GrAImes Expert group evaluated the microfictions written by expert writers with higher coherence and internal consistency.
			\begin{table}
				\scriptsize				
                \begin{minipage}{.45\linewidth}
                \caption{Intraclass Correlation Coefficient and Average Values of Literary Experts' Responses to GrAImes Questions.}\label{table:ICCAlpha_hwmf}
				\begin{tabular}{ c c c c} 
					\hline
					\multicolumn{4}{ c }{\textbf{Questions ICC - AVG}}  \\
					\hline				
					 Question & ICC &AV&SD\\ 
					\hline
					\textbf{3}&0.87& 3.5&1\\
					\hline
					\textbf{11}&0.75& 2.4 & 0.8\\
					\hline
					\textbf{10}&0.67& 3.6 & 1.7\\
					\hline
					\textbf{6}&0.65& 3.4 & 1.1\\
					\hline
					\textbf{5}&0.57& 3.2 & 1.4\\
					\hline
					\textbf{8}&0.55&1.8 & 0.9\\
					\hline
					\textbf{7}&0.29& 2.2&1.1\\ \hline
					\textbf{12}&0.21&2.2&1.1\\ \hline
					\textbf{9}&0.16&2.3&1.4\\ \hline
					\textbf{13}&-0.72&2&1.2\\ \hline
				\end{tabular} 
                \end{minipage}
                \hspace{.4cm}
                    \begin{minipage}{.45\linewidth}
                    \caption{Responses of Literary Experts to GrAImes Questions Evaluating Human Written Microfictions, Including Cronbach's Alpha for Internal Consistency, as well as AV and SD.}
				\begin{tabular}{ c c c c c} 
					\hline
					\multicolumn{5}{c}{\textbf{MF, Cronbanch's Alpha, Internal consistency , AV, SD}}  \\
					\hline				
					 MF& Alpha & IC& AV & SD\\ 
					\hline
					1& 0.8 &Good & 3 & 1.3\\
					\hline
					 2 & 0.79 & Acceptable &3.1 & 1.4\\
					\hline
					 4 & 0.75 & Acceptable & 2.5 & 1\\
					\hline
					 6 &0.67 & Questionable & 2.4 & 1.1\\
					\hline
					 3 &0.34 & Unacceptable & 2 & 0.9\\
					\hline
					 5 &0.13 & Unacceptable&2.9 & 0.9\\
					\hline
				\end{tabular}
                \end{minipage}
			\end{table}
			
                        Microfictions written by authors with medium experience (MF 4 and MF 6) displayed Alpha values of 0.75 and 0.67, respectively, which fall within the acceptable to questionable range. While the evaluations of these microfictions maintained moderate internal consistency, they exhibited higher standard deviations (SD = 1 and SD = 1.1) compared to the microfictions written by experts. This could imply that the expert evaluators are able to coherently distinguish between a good text from an emerging author and a less effective microfiction from a medium-experienced author.
    \begin{figure}[h]            
     \caption{Kendall W by GrAImes sections of the Expert group evaluation of human written microfictions.}\label{fig:kendall_w_all_sections}
     \vspace{.1cm}
        \includegraphics[width=.6\linewidth]{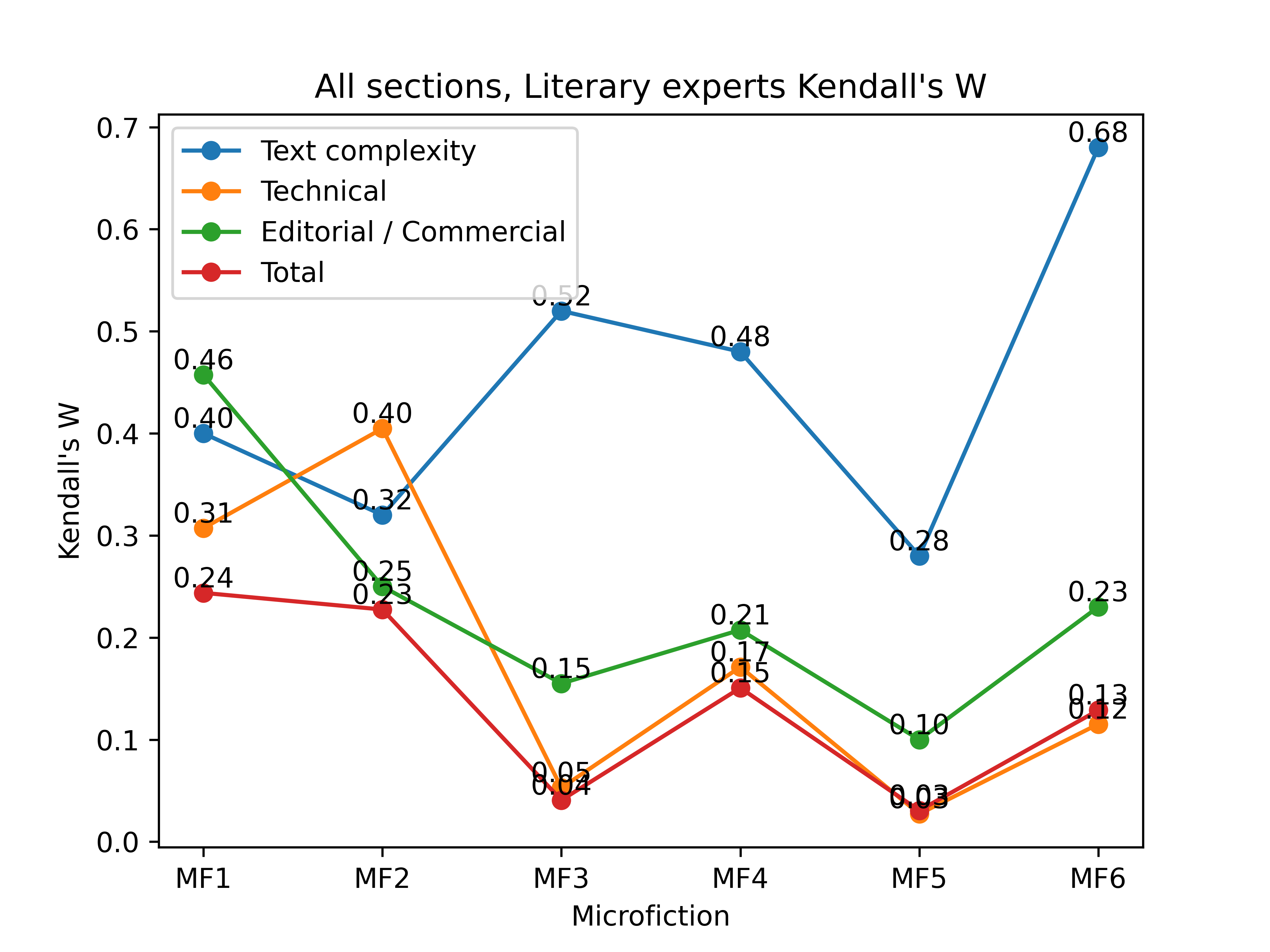}                   		\end{figure}

    Conversely, texts written by authors with low expertise (MF
    3 and MF 5) demonstrated the lowest internal consistency, with Alpha values of 0.34 and 0.13, respectively. These values are classified as unacceptable, suggesting significant inconsistencies within the text. The standard deviation (SD = 0.9 for both) was lower than that of the expert and medium-experience authors, which may indicate a lack of variability in linguistic structures or a rigid, less developed writing style. The low consistency of these texts highlights the challenges faced by less experienced authors in maintaining logical coherence and plot story structure.

    The results of the Kendall W (see Table \ref{fig:kendall_w_all_sections}) analysis indicate varying levels of agreement among experts, with the highest concordance observed in MF1 and MF2, which were written by an expert author. In contrast, MF3 and MF5 were authored by an emerging writter, showing lower levels of agreement. These findings suggest that an author's expertise aligns with the evaluations made by literary experts.

      Additionally, the average values of the microfictions provide further insight (see Table \ref{fig:lit_exps_beeswarm_hw_mfs}). Expert-authored texts had the highest AV (3 and 3.1), followed by medium-experience authors (2.5 and 2.4), while low-experience authors scored the lowest (2 and 2.9). This pattern reinforces the idea that writing expertise influences not only internal consistency but also the overall quality perception of the text.
		
		These findings align with existing research \citep{mccutchen2011novice} on the relationship between writing expertise and text coherence. Higher expertise leads to better-structured, logically consistent texts, whereas lower expertise results in fragmented, inconsistent writing. The judges rated microfictions written by a more experienced author higher and those written by a starting author lower. This is consistent with the purpose of our evaluation protocol, which aims to provide a tool for quantifying and qualifying a text based on its literary purpose as a microfiction.
		
		Among the evaluated questions (see table \ref{table:GrAImes_questions} Likert answer column), Question 3 exhibited the highest ICC (0.87), indicating excellent reliability and strong agreement among respondents. Its relatively high average score (AV = 3.5) and moderate standard deviation (SD = 1) suggest that participants consistently rated this question favorably. Similarly, Question 11 (ICC = 0.75) demonstrated good reliability, although its AV (2.4) was lower, suggesting that respondents agreed on a more moderate evaluation of the item, see table \ref{table:ICCAlpha_hwmf}.
		
		Moderate reliability was observed in Questions 10 and 6, with ICC values of 0.67 and 0.65, respectively. Their AV scores (3.6 and 3.4) suggest that they were generally well-rated, however, the higher standard deviation of Question 10 (SD = 1.7) indicates a greater spread of responses, possibly due to varying interpretations or differences in respondent perspectives. Questions 5 and 8, with ICC values of 0.57 and 0.55, respectively, fall into the questionable reliability range. Notably, Question 8 had the lowest AV (1.8), indicating that respondents found it more difficult or unclear, which may have contributed to the reduced agreement among responses.
		
		In contrast, Questions 7, 12, and 9 exhibited low ICC values (0.29, 0.21, and 0.16, respectively), suggesting weak reliability and higher response variability. The AV values for these items ranged from 2.2 to 2.3, further indicating inconsistent interpretations among participants. The standard deviations for these questions (SD = 1.1–1.4) suggest a broad range of opinions, reinforcing the need for potential revisions to improve clarity and consistency.
		
		A particularly notable finding is the negative ICC value for Question 13 (-0.72). Negative ICC values typically indicate systematic inconsistencies, which may stem from ambiguous wording, multiple interpretations, or flaws in question design. With an AV of 2.0 and an SD of 1.2, it is evident that responses to this item lacked coherence. 
		
\begin{figure}[h!]
            \caption{Interrater agreement on clarity (q1), structure(q2), complexity (q4), giftability (q14), and commerciality (q15) for microfiction 2 from the Expert group}
			\label{fig:lit_exp_humw_microfiction2question1}
			\vspace{.2 cm}
			\includegraphics[width=.5 \linewidth]{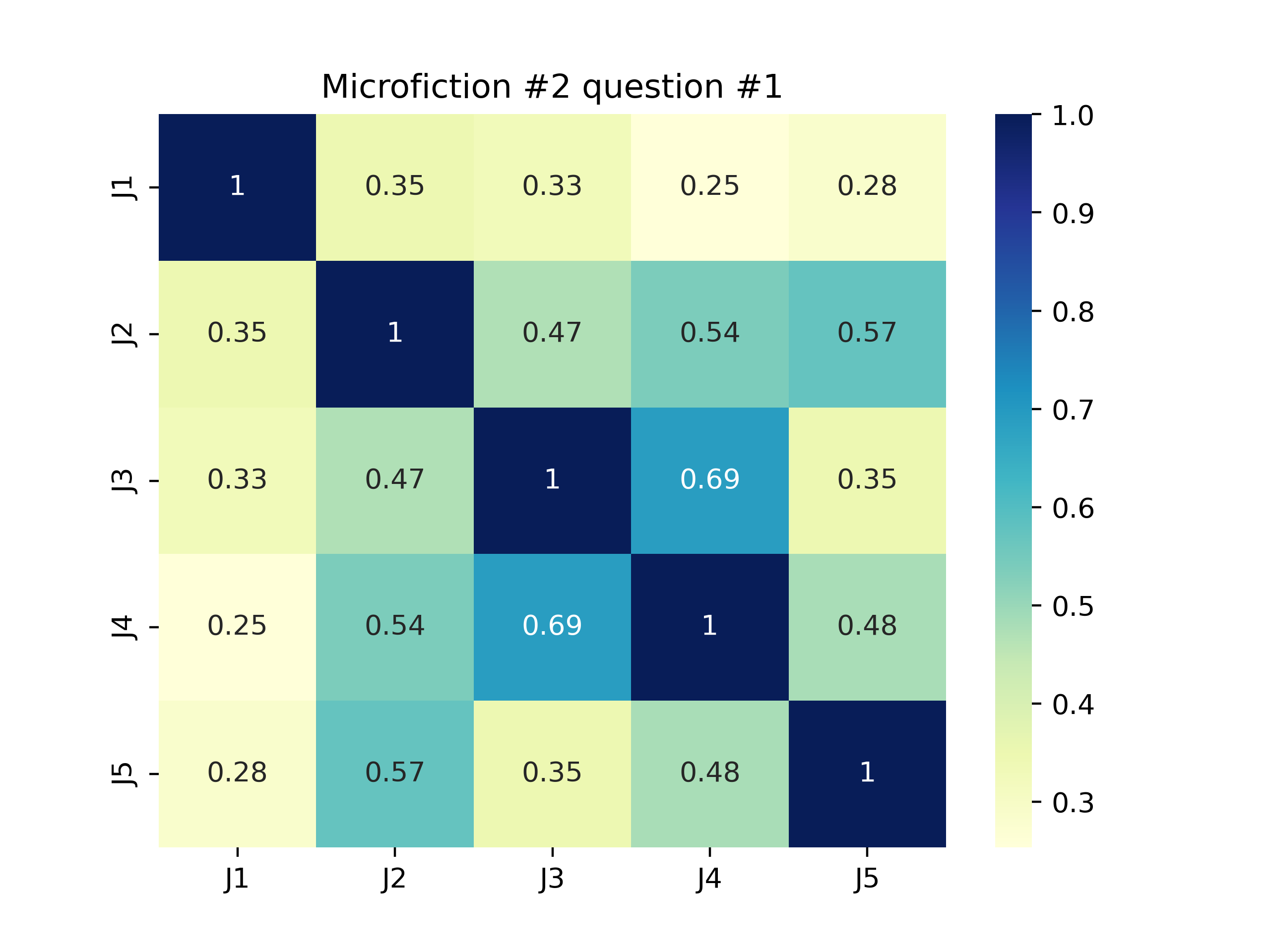}
            \includegraphics[width=.5 \linewidth]{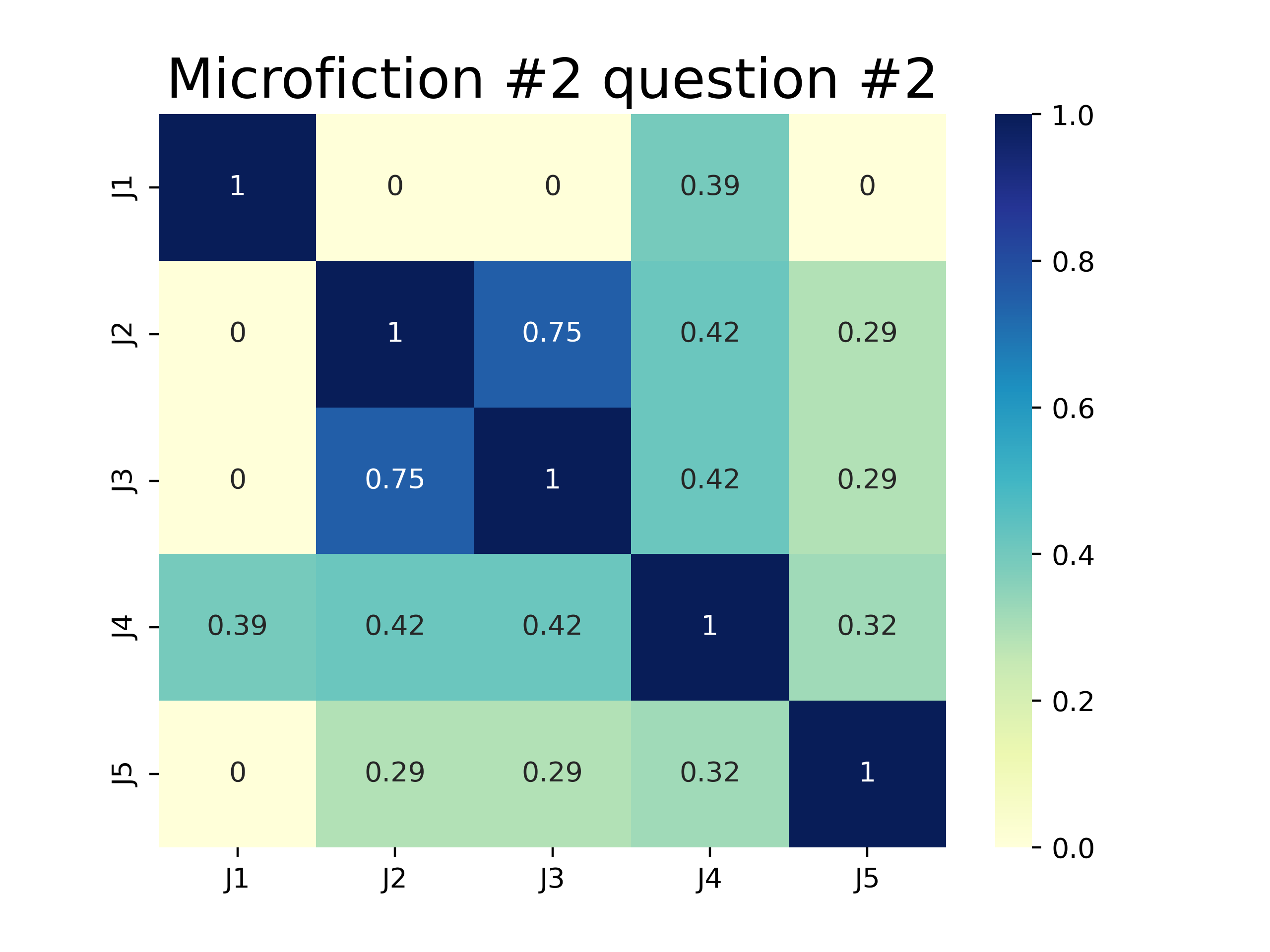}
		
		\end{figure}

		Regarding the responses to the 5 open answer questions (see numbers 1, 2, 4, 14, and 15 in table \ref{table:GrAImes_questions}) we used Sentence-BERT \citep{reimers2019sentence}, and semantic cosine similarity \citep{rahutomo2012semantic} to look for lexical and semantic similarities between GrAImes open answer questions. These results reveal key insights into evaluation agreement and interpretation variability across the six microfictions. For Question 1 (plot comprehension), agreement was often weak (e.g., J1-J4 semantic cosine similarity = 0.21 in microfiction 1), suggesting narrative ambiguity or divergent reader focus. Question 2 (theme identification) showed inconsistent alignment (e.g., J2-J3 similarity = 0.67 in microfiction 2 vs. J1-J3 = 0.10 in microfiction 1, see figure \ref{fig:lit_exp_humw_microfiction2question1}), indicating subjective thematic interpretation. Question 4 (interpretation specificity) had polarized responses, with perfect agreement in some cases (e.g., J1-J2 = 1.00 in microfiction 3) but stark divergence in others (J2-J3 = 0.00 in microfiction 4), reflecting conceptual or terminological disparities. Questions 14 (gifting suitability) and 15 (publisher alignment) demonstrated higher consensus (e.g., perfect agreement among four judges in microfiction 4, Q4), likely due to more objective criteria. However, J5 consistently emerged as an outlier (e.g., similarity $\leq$ 0.11 in microfiction 1, Question 15), underscoring individual bias. The protocol’s value lies in quantifying these disparities: clearer questions (14-55) reduced variability, while open-ended ones (1-2) highlighted the need for structured guidelines to mitigate judge-dependent subjectivity, particularly in ambiguous or complex microfictions.

		On the two extra questions given to the literary experts (see \ref{two_extra_questions}), the majority of experts (3 out of 5) found the microfiction evaluation protocol sufficiently clear for use, while a minority (2 out of 5) expressed concerns regarding ambiguous or unclear criteria. 
		A strong consensus (4 out of 5 experts) agreed that the protocol can effectively evaluate the literary value of microfiction. However, one dissenting opinion highlights the need for adjustments in specific criteria to ensure a more precise assessment.
		
		\subsection{GrAImes evaluation of Monterroso and ChatGPT-3.5 generated microfictions, evaluated by the Enthusiast group}
 		GrAImes was applied to assess a collection of six microfictions crafted by advanced AI tools. These tools include the renowned short story creator inspired by the style of renowned Guatemalan author Augusto Monterroso and ChatGPT-3.5. The literature enthusiasts who participated in this study evaluated the microfictions based on parameters such as coherence, thematic depth, stylistic originality, and emotional resonance.	 
				
		A total of six microfictions were generated, with three created by the Monterroso tool (MF = 1, 2, 3) and three by ChatGPT-3.5 (MF = 4, 5, 6). The microfictions were evaluated on a Likert scale of 1 to 5, with ratings provided by a panel of 16 reader enthusiasts. The average and standard deviation (SD) of ratings were calculated for each microfiction. The results of the analysis are presented in table \ref{table:Lit_enth_resps_all}.

            \begin{figure}
                \centering
                \caption{Enthusiast group evaluation of AI-generated microfictions}
                \vspace{.2 cm}                \label{fig:lit_enth_beeswarm_ai_mfs}                    \includegraphics[width=0.4\linewidth]{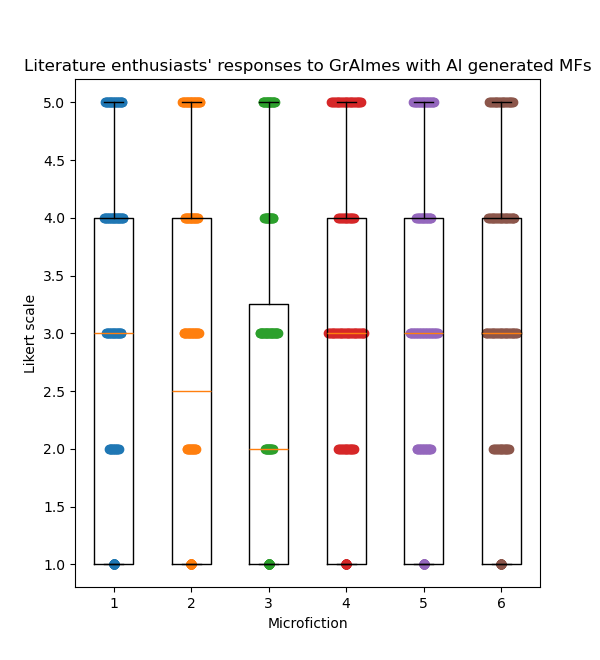} 
            \end{figure}

            \begin{figure}
                \caption{nthusiast group evaluation of AI-generated microfictions by section.}
                \vspace{.2 cm}
                \centering
                \label{fig:lit_enth_bargrphs_ai_mfs}       \includegraphics[width=0.4\linewidth]{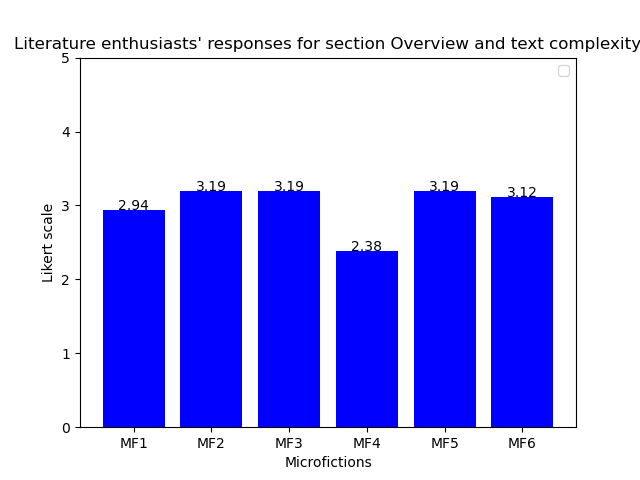}
            \includegraphics[width=0.4\linewidth]{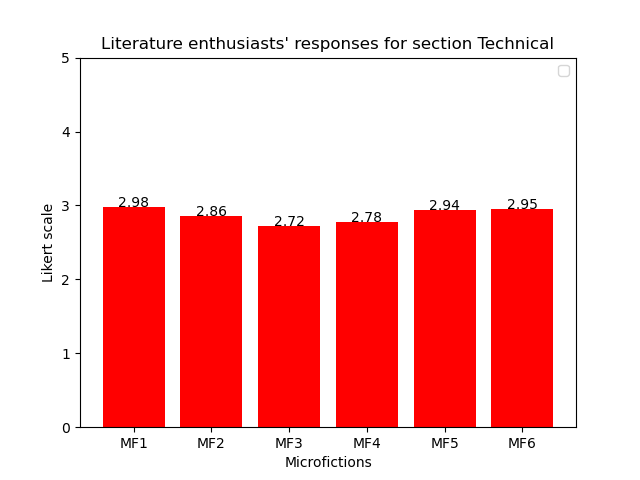}
            \includegraphics[width=0.4\linewidth]{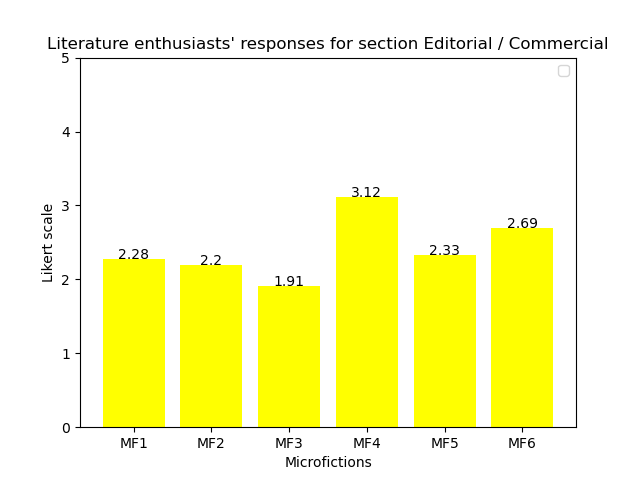}
            \includegraphics[width=0.4\linewidth]{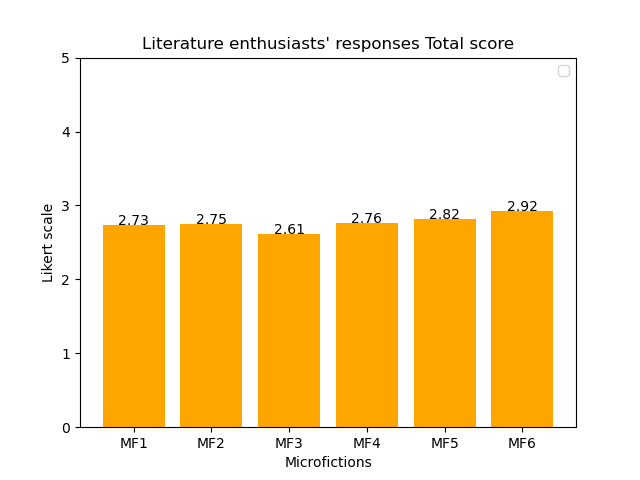}
            \end{figure}

		\begin{table}[h]
			\tiny	
			\caption{Responses, AV and SD, of the Enthusiast group  to AI-generated microfictions (MF1-MF6), measured on a Likert Scale and Grouped by GrAImes Questions Sections, with a Total Average Responses Final Column.}
			\label{table:Lit_enth_resps_all}			
			\begin{tabular}{
					m{35em}  m{.5em} m{.5em}
					m{.5em} m{.5em}  m{.5em} m{.5em} 
					m{.5em} m{.5em}  m{.5em} m{.5em}
					m{.5em} m{.5em}  m{.5em} m{1em}
				} 	
				\hline
				\multicolumn{15}{c}{\textbf{Literature enthusiasts responses to Microfictions from Monterroso and ChatGPT-3.5}}\\
				\hline
				& \multicolumn{2}{c}{MF 1} & \multicolumn{2}{c}{MF 2}& \multicolumn{2}{c}{MF 3} & \multicolumn{2}{c}{MF 4}& \multicolumn{2}{c}{MF 5} & \multicolumn{2}{c}{MF 6} & \multicolumn{2}{c}{Average}\\
				\hline
				Question & AV & SD & AV & SD & AV & SD  & AV & SD & AV & SD& AV & SD & AV & SD \\
				\hline
				\multicolumn{15}{c}{\textbf{Story Overview and text complexity}} \\
				\hline
				3.-Does it propose other interpretations, in addition to the literal one? & 2.9  & 1.5 & 3.2 & 1.6 & 2.9 & 1.7& 2.4 & 1.5 & 3.2 & 1.6 & 3.1 & 1.6& 2.9 & 1.6\\ 
				\hline
				\multicolumn{15}{c}{\textbf{Technical}} \\
				\hline
				5.-Is the story credible?& 1.9 &0.9  & 1.7 &0.9 &2.2 &1.1 &4.2 &1.2  &4  &1.2  &4.3 &0.9 &3.1 & 1\\ 
				\hline
				6.-Does the text require your participation or cooperation to complete its form and meaning? & 4.6 & 1  & 4.3 & 1.4 & 4.3 & 1. & 2.4 & 1.2 & 3.1 &1.4 & 2.9 & 1.4 & 3.6 & 1.3 \\
				\hline
				7.-Does it propose a new vision of reality?  & 2.7 &1.7  & 2.9& 1.5 & 2.4 & 1.5 & 2.3 & 1.4 & 2.7 & 1.3 & 2.5 &1.3 & 2.6& 1.4\\
				\hline
				8.-Does it propose a new vision of the genre it uses? & 2.3  & 1.4 & 2.7 & 1.6 & 2.1 & 1.3 & 2.4 & 1.5 & 2.4 & 1.5 & 2.4 & 1.1& 2.4 &1.4 \\  
				\hline
				9.-Does it propose a new vision of the language itself? & 3.4 & 1.3 & 2.7 & 1.4 & 2.6 & 1.3 & 2.6 & 1.3 & 2.4 & 1.4 & 2.7 & 1.3 & 2.7 & 1.3\\  
				\hline
				\multicolumn{15}{c}{\textbf{Editorial / commercial}}  \\
				\hline
				10.-Does it remind you of another text or book you have read? & 2.9 & 1.5 &  2.8 & 1.3 & 2.9 & 1.5 & 3.9 & 1.3 & 3.2 & 1.5 & 3.2& 1.5 & 3.2 & 1.4\\  
				\hline
				11.-Would you like to read more texts like this?  & 2 & 1.2 & 2.3 & 1.7 & 1.7 & 0.9 & 3 & 1.5 & 2.1 & 1.3 & 2.6 & 1.5 & 2.3 & 1.4 \\ 
				\hline 
				12.-Would you recommend it? & 2.1 & 1.5 & 2.1 & 1.5 & 1.6 & 0.9 & 2.8 & 1.4 & 2.1 & 1.4 & 2.6 & 1.6 & 2.2 & 1.4\\
				\hline
				13.-Would you give it as a present? & 2.1 & 1.6 & 1.7 & 1.3 & 1.4 & 1 & 2.8 & 1.5 & 2 & 1.4 & 2.3 & 1.4 & 2.1 & 1.4 \\ 
				\hline
			\end{tabular}
		\end{table}
				
\begin{figure}[h]
			\caption{Line charts of literature enthusiasts GrAImes sections summarized AV and SD}
			\label{fig:line_chart_lit_ent_graimes_sections}
			\vspace{.2 cm}
			\includegraphics[width=7cm]{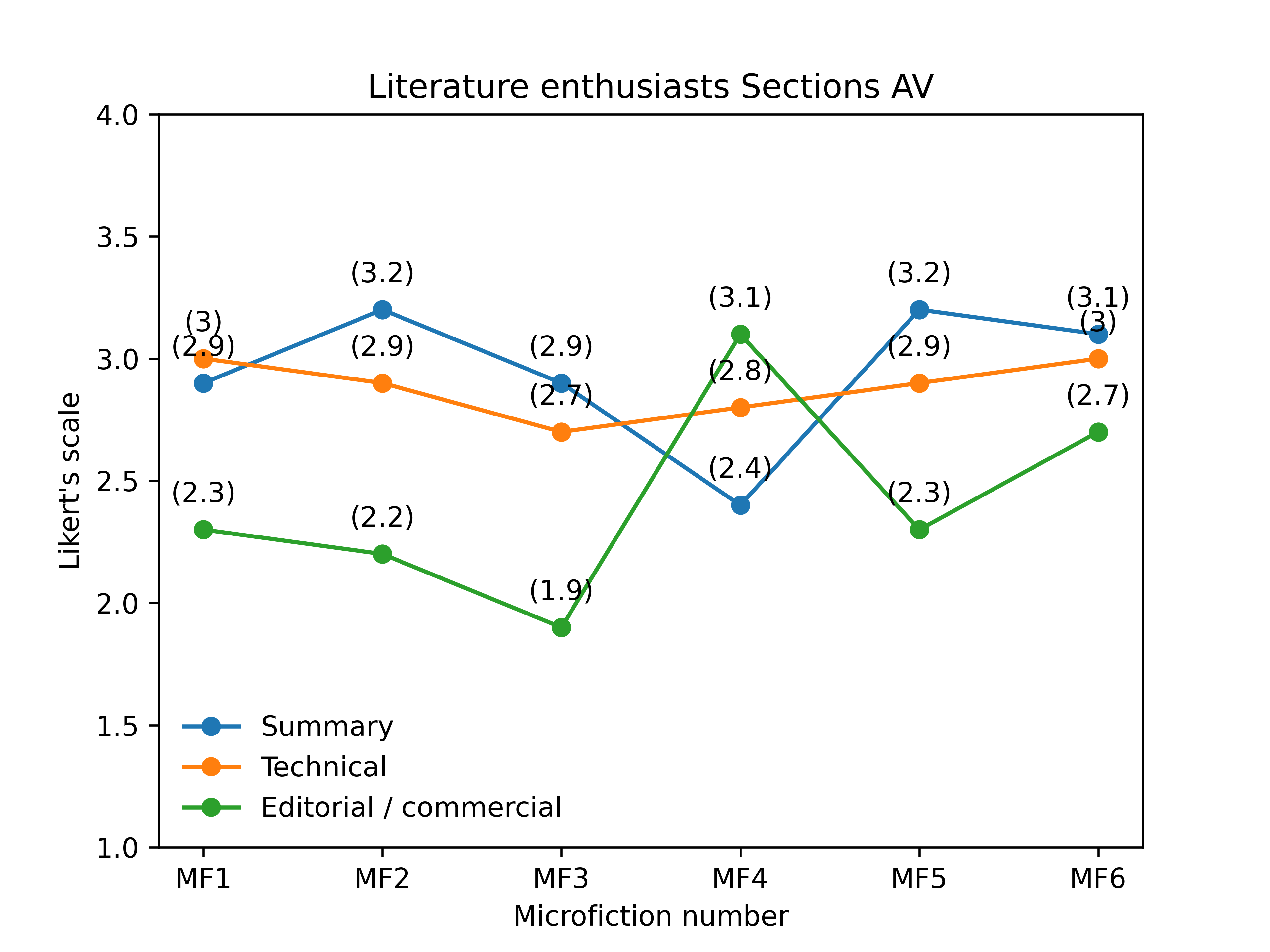}
			\includegraphics[width=7cm]{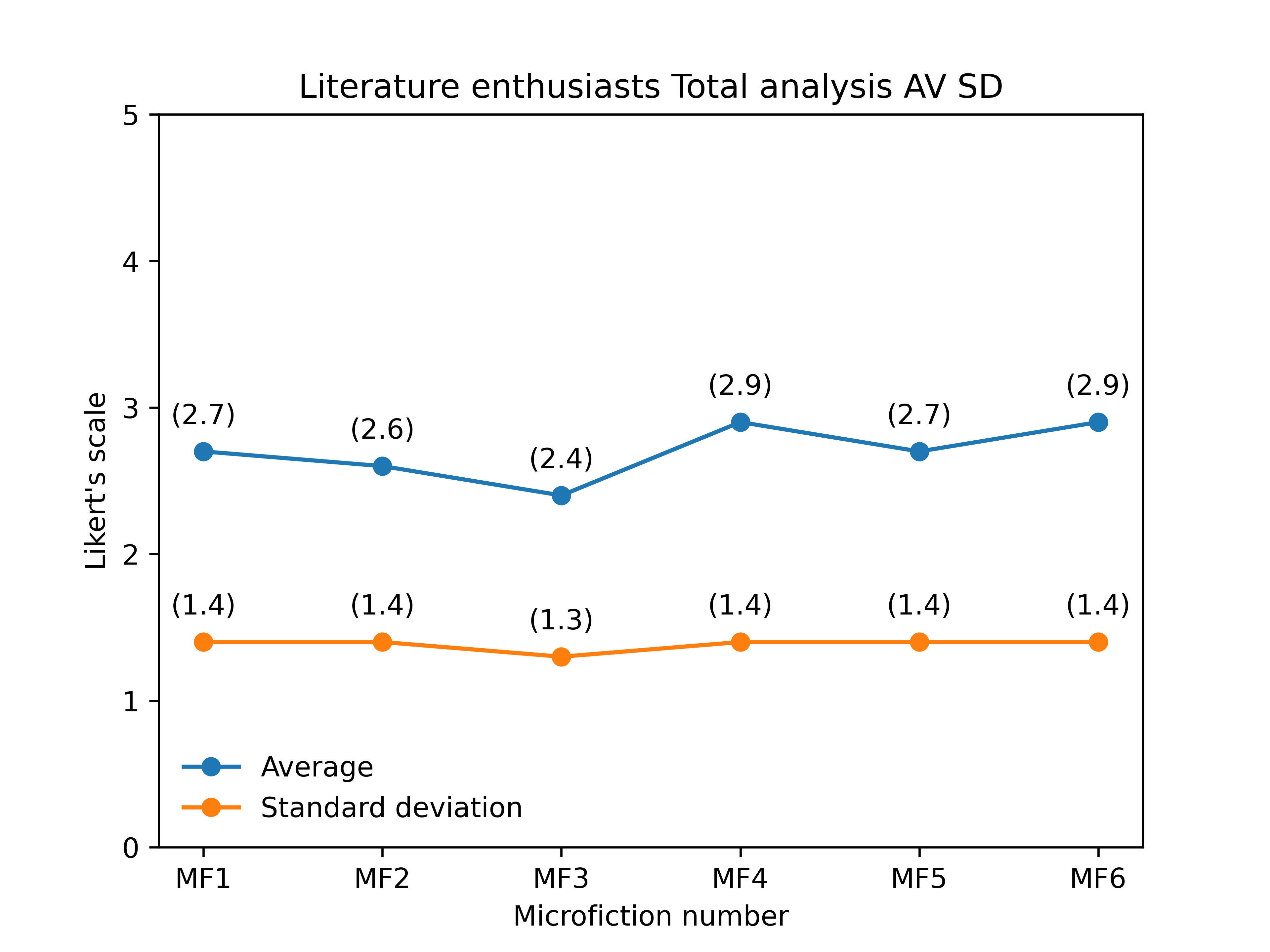}
		\end{figure}

		The results indicate that the ChatGPT-3.5 microfictions (MF = 4, 5, 6) have slightly higher average ratings (2.7-2.9) compared to the Monterroso-generated microfictions (MF = 1, 2, 3), which have average ratings ranging from 2.4 to 2.7 (see tables \ref{table:Lit_enth_sections_AV_SD} and figures \ref{fig:line_chart_lit_ent_graimes_sections}). The standard deviation values are consistent across most microfictions, indicating a relatively narrow range of ratings.
		
		\begin{table}[h]
			
                \scriptsize
                \label{table:Lit_enth_sections_AV_SD}
		\begin{minipage}{.25\linewidth}
                \caption{Section Story Overview and text complexity AV and SD by MF}
				
			\begin{tabular}{c c c c } 
				\hline
				\multicolumn{4}{c}{\textbf{Overview and Complexity}}  \\
				\hline
				\#& \textbf{MF} & \textbf{AV} & \textbf{SD}\\ 
				\hline
				1& 2 & 3.2 & 1.6\\
				\hline
				2&5 & 3.2 & 1.6\\
				\hline
				3& 6 & 3.1 & 1.6\\
				\hline
				4& 3 & 2.9 & 1.7\\
				\hline
				5& 1 & 2.9 & 1.5\\
				\hline
				6& 4 & 2.4 & 1.5\\
				\hline
			\end{tabular}
		\end{minipage}	
        \hspace{.2cm}
        		\begin{minipage}{.2\linewidth}
                \caption{Section Technical AV and SD by MF}
            \begin{tabular}{c c c}
				\hline
				\multicolumn{3}{c}{\textbf{Technical}}  \\ 
				\hline
				\textbf{MF} & \textbf{AV} & \textbf{SD}\\  
				\hline\hline
				6 & 3 & 1.2\\
				\hline
				1 & 3 & 1.3\\
				\hline
				2 & 2.9 & 1.4\\
				\hline
				5 & 2.9 & 1.3\\
				\hline
				4 & 2.8 & 1.3\\
				\hline
				3 & 2.7 & 1.3\\
				\hline
			\end{tabular}
                \end{minipage}
            \hspace{.2cm}
            \begin{minipage}{.25\linewidth}
                \caption{Section Editorial \ Commercial AV and SD by MF}
			\begin{tabular}{c c c} 
				\hline
				\multicolumn{3}{c}{\textbf{Editorial/commercial}}  \\
				\hline
				\textbf{MF} & \textbf{AV} & \textbf{SD}\\  
				\hline\hline
				4 & 3.1 & 1.5\\
				\hline
				6 & 2.7& 1.5\\
				\hline
				1 & 2.3 & 1.4\\
				\hline
				5 & 2.3 & 1.4\\
				\hline
				2 & 2.2 & 1.4\\
				\hline
				3 & 1.9 & 1.1\\
				\hline
			\end{tabular}
                \end{minipage}
            \hspace{.4cm}
            \begin{minipage}{.2\linewidth}
                \caption{Total Analysis AV and SD by MF}
			\begin{tabular}{c c c} 
				\hline
				\multicolumn{3}{c}{\textbf{Total Analysis}}  \\
				\hline				
				\textbf{MF} & \textbf{AV} & \textbf{SD}\\ 
				\hline\hline
				4 & 2.9 & 1.4\\
				\hline
				6& 2.9 & 1.4\\
				\hline
				5 & 2.7 & 1.4\\
				\hline
				1 & 2.7 & 1.4\\
				\hline
				2 & 2.6 &1.4\\
				\hline
				3 & 2.4 & 1.3\\
				\hline
			\end{tabular}
            \end{minipage}
		\end{table}
		
		The most consistent response pertains to the credibility of the stories (AV = 3.1, SD = 1.0), indicating a strong agreement among participants that the narratives were believable. This suggests that, regardless of other literary attributes, the microfictions maintain a sense of realism that resonates with readers. The question regarding whether the text requires the reader's participation or cooperation to complete its form and meaning received the highest average rating (AV = 3.6, SD = 1.3). This suggests that the microfictions engage readers actively, requiring interpretation and involvement to fully grasp their meaning. The relatively low SD indicates moderate consensus on this aspect.
		
		Questions concerning literary innovation—whether the texts propose a new vision of language (AV = 2.7, SD = 1.3), reality (AV = 2.6, SD = 1.4), or genre (AV = 2.4, SD = 1.4)—show moderate variation in responses. This suggests that while some readers perceive novelty in these areas, others do not find the texts particularly innovative. Similarly, the question of whether the texts remind readers of other books (AV = 3.2, SD = 1.4) presents a comparable level of divergence in opinions. The lowest-rated questions relate to the desire to read more texts of this nature (AV = 2.3), the willingness to recommend them (AV = 2.2), and the inclination to gift them to others (AV = 2.1), all with SD = 1.4. These results suggest that while the microfictions may have some engaging qualities, they do not strongly motivate further exploration or endorsement.
		
		Interestingly, the question about whether the texts propose interpretations beyond the literal one received the highest standard deviation (SD = 1.6, AV = 2.9). This indicates significant variation in responses, suggesting that some readers found deeper layers of meaning, while others perceived the texts as more straightforward.
                
		The Intraclass Correlation Coefficient analysis of GrAImes answers (see table \ref{table:lit_enth_ICC_Alpha}) revealed varying degrees of reliability among the 16 literature enthusiasts raters when assessing texts generated by Monterroso and ChatGPT-3.5. Three questions demonstrated poor reliability (ICC \textless 0.50), reflecting high variability in responses, with Question 8 exhibiting a negative ICC (-0.44), suggesting severe inconsistency, possibly due to misinterpretation or extreme subjectivity. In contrast, Questions 5 and 6 showed excellent reliability (ICC \textgreater 0.90), indicating strong inter-rater agreement, while Questions 9, 11, 12, and 13 displayed moderate reliability (ICC 0.60–0.70), implying acceptable but inconsistent consensus. These findings highlight the need to refine ambiguous or subjective questions to improve evaluative consistency in microfiction assessment.
		
		\begin{table}[h]	
			\scriptsize
			\caption{Internal Consistency Analysis: ICC and Cronbach’s Alpha for Microfictions from Monterroso and ChatGPT-3.5 Evaluated by Literarture Enthusiasts.}
			\label{table:lit_enth_ICC_Alpha}
			
			\begin{tabular}{c c c c c } 
				\hline
				\multicolumn{5}{c}{\textbf{Literature enthusiasts ICC - AVG - SD Analysis}}  \\
				\hline				
				\#& Question & ICC &AV&SD\\  
				\hline
				1&5&0.97& 3.1 & 1\\
				\hline
				2& 6 &0.95& 3.6&1.3\\\hline
				3& 13 &0.70&2.1&1.4\\ \hline
				4& 9 &0.67& 2.7 & 1.3\\\hline
				5& 11 &0.67&2.3&1.4\\ \hline
				6& 12 &0.62&2.3&1.4\\ \hline
				7& 10 &0.57& 3.2 & 1.4\\\hline
				8& 3 &0.28& 2.9 & 1.6\\\hline
				9& 7 &0.01& 2.6 & 1.4\\\hline
				10& 8 &-0.44& 2.4&1.4\\ \hline	
			\end{tabular} \hspace{.5 cm}
			\begin{tabular}{c c c c c c} 
				\hline
				\multicolumn{6}{ c}{\textbf{MF, Alpha, Internal consistency (IC), AV, SD}}  \\
				\hline				
				\#& MF& Alpha & IC & AV & SD\\ 
				\hline\hline
				1& 4 & 0.90 & Excellent & 2.9 & 1.4\\
				\hline
				2& 5 & 0.89 & Good & 2.7 & 1.4\\
				\hline
				3& 6 & 0.89 & Good & 2.9 & 1.4\\
				\hline
				4& 1 &0.88 & Good & 2.7 & 1.4\\
				\hline
				5& 2 &0.84 & Good & 2.6 & 1.4\\
				\hline
				6& 3 & 0.79 & Acceptable& 2.4 & 1.3\\
				\hline
			\end{tabular}
		\end{table}
		
		\begin{figure}[]
			\caption{ICC line chart, literature enthusiasts responses to AI generated microfictions.}
			\vspace{.2 cm}
			\includegraphics[width=.5\linewidth]{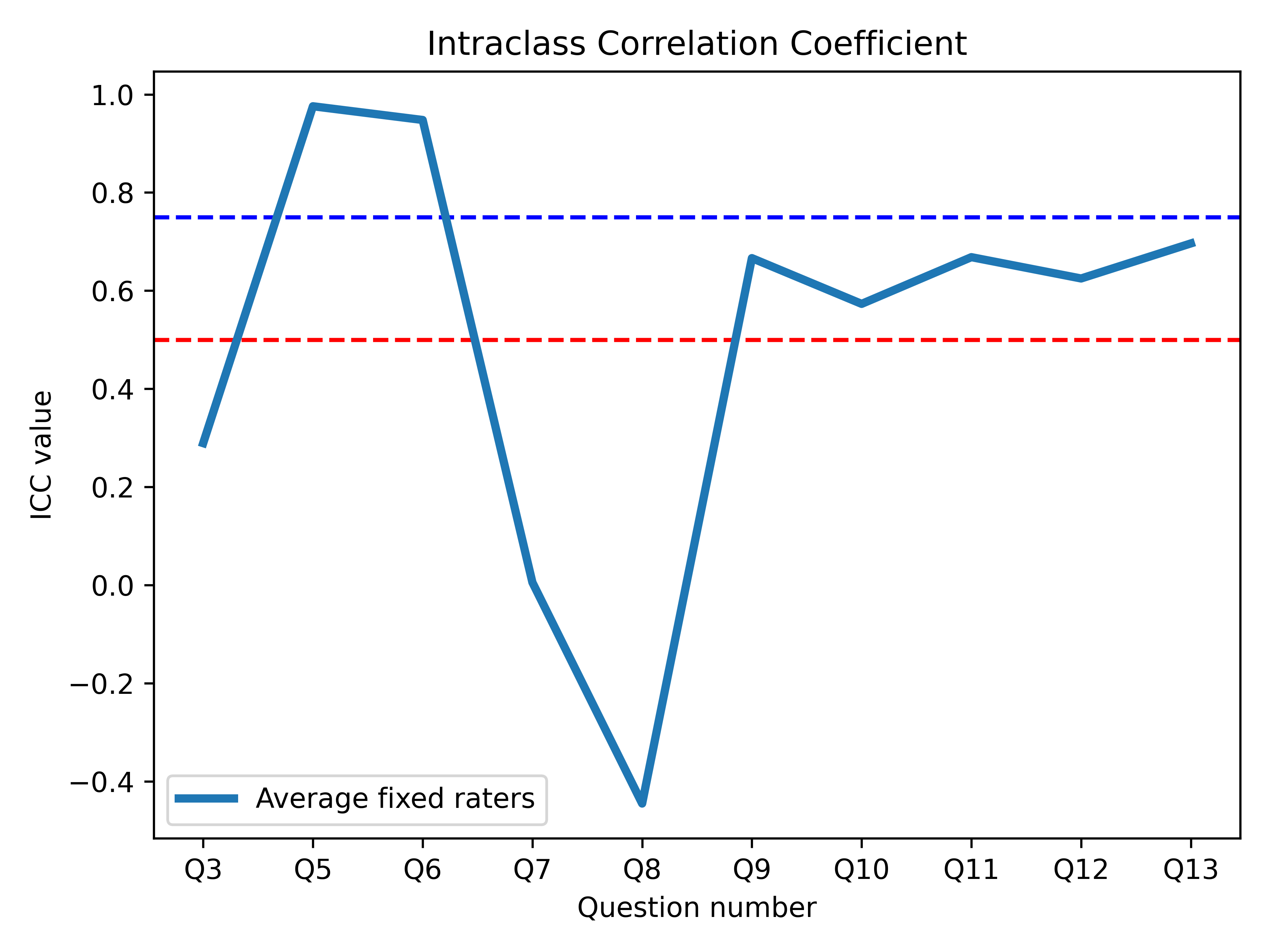}
		\end{figure}
		

        The study assessed microfictions (MFs) based on their ability to propose interpretations beyond the literal meaning and found notable differences between Monterroso's (MF 1–3) and ChatGPT-3.5's (MF 4–6) texts. Monterroso's MF 2 had the highest average score (AV = 3.2), showing a stronger ability to suggest multiple interpretations, while ChatGPT-3.5's MF 4 had the lowest score (AV = 2.4), indicating limited interpretive depth. Standard deviation values were consistent across all MFs (1.5 to 1.7), showing moderate response variability among literature enthusiasts. Thus, while some MFs were seen as more interpretively rich the response variability was similar for all texts.

        The technical quality of the MFs was assessed through questions related to credibility (Question 5), reader participation (Question 6), and innovation in reality, genre, and language (Questions 7–9). MF 6, generated by the ChatGPT-3.5, scored highest in credibility (AV = 4.3), while MF 1, generated by Monterroso, scored the lowest (AV = 1.9). This indicates a clear distinction in perceived realism between the two sources, as evaluated by literature enthusiasts. In terms of reader participation, MF 1 scored highest (AV = 4.6), suggesting it effectively engaged readers in completing its form and meaning. However, MF 4 scored the lowest in this category (AV = 2.4), highlighting a potential weakness in ChatGPT-3.5's generated texts. Innovation in language (Question 9) was highest in MF 1 (AV = 3.4), while MF 5 scored the lowest (AV = 2.4). Overall, the technical quality of MFs generated by Monterroso (MF 1–3) was slightly higher (AV = 2.7–3.0) compared to those generated by the ChatGPT-3.5 (AV = 2.8–3.0), with MF 3 scoring the lowest (AV = 2.7). The consistent SD values (ranging from 0.9 to 1.7) reflect similar levels of variability in responses from literature enthusiasts.
		
		The editorial and commercial appeal of the MFs was evaluated based on their resemblance to other texts (Question 10), reader interest in similar texts (Question 11), and willingness to recommend or gift the texts (Questions 12–13). MF 4, generated by the ChatGPT-3.5, scored highest in resemblance to other texts (AV = 3.9), while MF 2 scored the lowest (AV = 2.8). This suggests that ChatGPT-3.5 generated texts may be more reminiscent of existing literature, as perceived by literature enthusiasts. In terms of reader interest, MF 4 also scored the highest (AV = 3.0), while MF 3 scored the lowest (AV = 1.7). Similarly, MF 4 was the most recommended (AV = 2.8) and most likely to be gifted (AV = 2.8), indicating stronger commercial appeal compared to Monterroso's generated texts. Overall, the ChatGPT-3.5's generated MFs (MF 4–6) outperformed Monterroso's generated MFs (MF 1–3) in editorial and commercial appeal, with MF 4 achieving the highest average score (AV = 3.1) and MF 3 the lowest (AV = 1.9). The SD values (ranging from 0.9 to 1.7) indicate moderate variability in responses from literature enthusiasts.
		
		The total analysis of the MFs (see table \ref{table:lit_exp_mont_chatgpt_summ_psect}) reveals that ChatGPT-3.5 texts (MF 4–6) generally outperformed Monterroso's generated texts (MF 1–3) in terms of editorial and commercial appeal, while Monterroso's texts showed slightly better technical quality. MF 4, generated by the ChatGPT-3.5, achieved the highest overall score (AV = 2.9), while MF 3, generated by Program A, scored the lowest (AV = 2.4). The standard deviation values were consistent across all categories (SD $\approx$ 1.3–1.4), indicating similar levels of variability in responses from literature enthusiasts. These findings suggest that while ChatGPT-3.5 texts may have stronger commercial potential, Monterroso's texts exhibit slightly higher technical sophistication. The evaluation by literature enthusiasts provides valuable insights into how general audiences perceive and engage with these microfictions.
				
		The evaluation of six microfictions by the Enthusiast group leaders revealed notable differences between microfictions generated by Monterroso (MF 1–3) and those generated by ChatGPT-3.5 (MF 4–6), see table \ref{table:lit_expt_montchat_all}. In terms of Story Overview and text complexity, MF 4 and MF 5 scored highest (AV = 4) for proposing multiple interpretations (see table \ref{table:lit_exp_mont_chatgpt_summ_psect} and figure \ref{fig:line_chart_lit_exp_graimes_sections}), while MF 3 scored the lowest (AV = 1, SD = 0), indicating a lack of depth. The Technical aspects showed that MF 1 and MF 4–6 were rated highly for credibility (AV = 5, SD = 0), whereas MF 3 scored poorly (AV = 2, SD = 1.4). MF 1 and MF 5 excelled in requiring reader participation (AV = 5, SD = 0), while MF 4 and MF 6 scored lower (AV = 3.5, SD = 0.7). However, all microfictions struggled to propose new visions of reality, language, or genre, with most scores ranging between 1 and 2.
	\begin{figure}
                \centering
                \caption{Enthusiast group evaluation of AI-generated microfictions.}
                \vspace{.2 cm}
                \label{fig:lit_exps_beeswarm_ai_mfs}\includegraphics[width=0.5\linewidth]{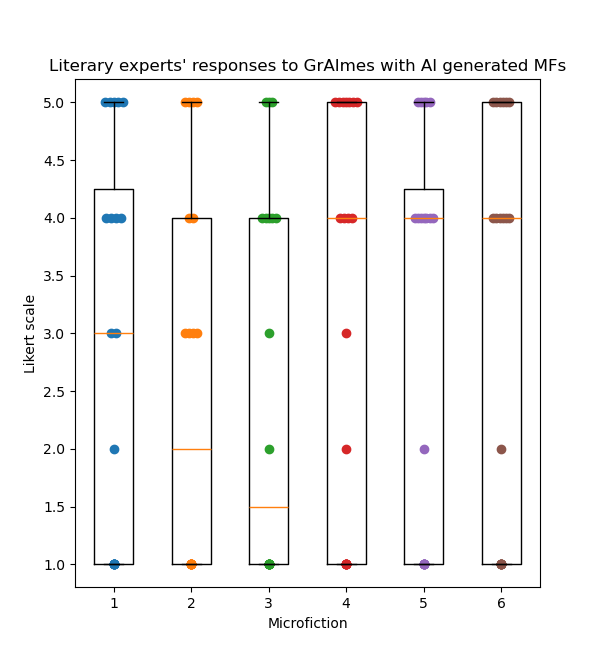}
        \end{figure}
        \begin{figure}
            \centering
                \caption{Enthusiast group evlauation of AI-generated microfictions (average by section).}
                \label{fig:lit_esps_bargrphs_ai_mfs}
                \vspace{.2 cm}
            
            \includegraphics[width=0.4\linewidth]{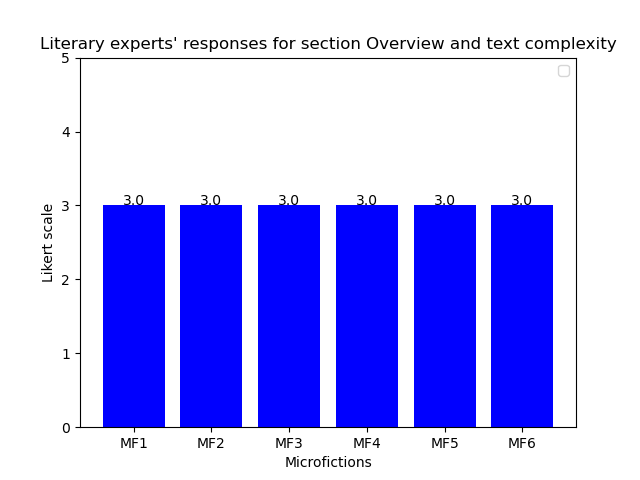}
                \includegraphics[width=0.4\linewidth]{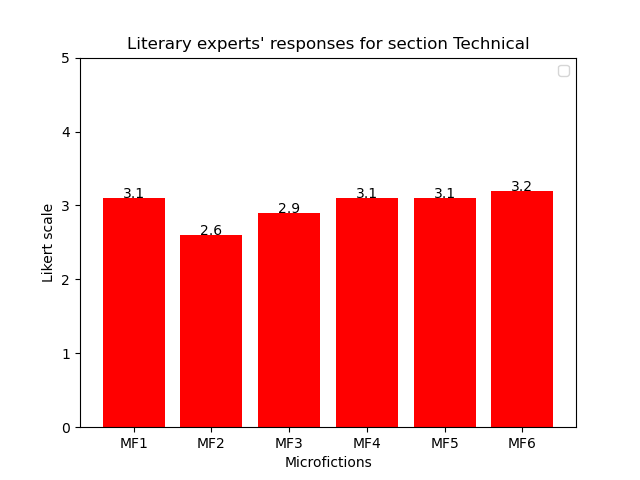}
                \includegraphics[width=0.4\linewidth]{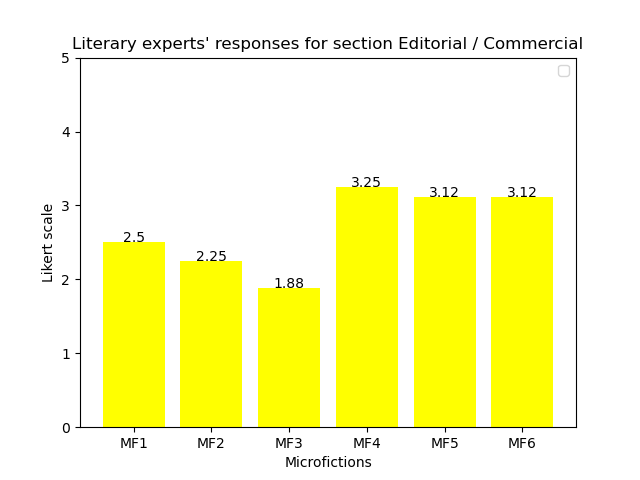}
                \includegraphics[width=0.4\linewidth]{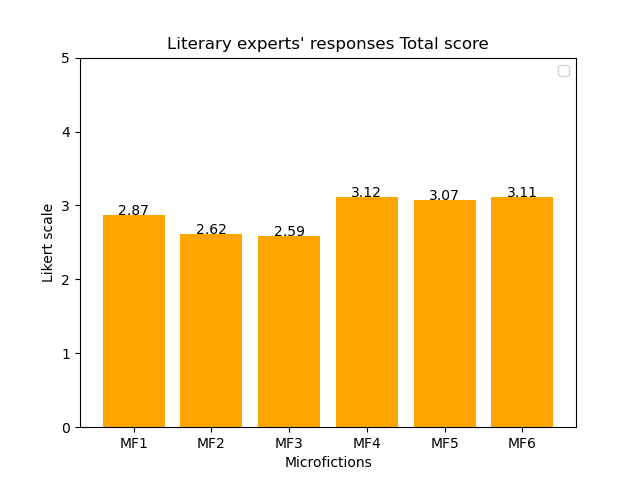}

            \end{figure}

		In the Editorial/Commercial category, MF 4–6 outperformed MF 1–3. MF 4 and MF 6 were most reminiscent of other texts (AV = 5 and 4.5, respectively) and were more likely to be recommended or given as presents (AV = 4, SD = 1.4). In contrast, MF 1–3 scored poorly in these areas, with MF 3 consistently receiving the lowest ratings (AV = 1, SD = 0). Overall, the ChatGPT-3.5 microfictions (MF 4–6) achieved higher total scores (AV = 3.4, SD = 0.8) compared to those generated by Monterroso (MF 1–3, AV = 2.2, SD = 1.1).
		
		\begin{table}[h]
			\tiny		
			\caption{Average Values (AV) and Standard Deviations (SD) of Literary Experts’ Responses to Microfictions by Monterroso and ChatGPT-3.5} 
			\label{table:lit_expt_montchat_all}	
			\begin{tabular}{
					m{35em} m{.5em} m{.5em}
					m{.5em} m{.5em} m{.5em} m{.5em} 
					m{.5em} m{.5em} m{.5em} m{.5em}
					m{.5em} m{.5em} m{.5em} m{1em}
				} 	
				\hline
				\multicolumn{15}{c}{\textbf{Literary experts responses to Microfictions from Monterroso and ChatGPT-3.5}}\\
				\hline
				& \multicolumn{2}{c}\textbf{MF 1} & \multicolumn{2}{c}\textbf{MF 2}& \multicolumn{2}{c}\textbf{MF 3} & \multicolumn{2}{c}\textbf{MF 1}& \multicolumn{2}{c}\textbf{MF 5} & \multicolumn{2}{c}\textbf{MF 6} & \multicolumn{2}{c}{Average}\\
				\hline
				Question & AV & SD & AV & SD & AV & SD  & AV & SD & AV & SD& AV & SD & AV & SD \\
				\hline
				\multicolumn{15}{c} Story Overview and text complexity \\
				\hline
				3.-Does it propose other interpretations, in addition to the literal one? & 3 & 2.8 & 3 & 2.8 & 1 & 0 & 4 & 1.4 & 4  & 1.4 & 3.5 & 2.1 & 3.1 & 1.8 \\ 
				\hline
				\multicolumn{15}{c}{\textbf{Technical}} \\
				\hline
				5.-Is the story credible?& 5 & 0 & 3 & 2.8 & 2 & 1.4 & 5 & 0 & 5 & 0 & 5 & 0 & 4.2 & 0.7\\ 
				\hline
				6.-Does the text require your participation or cooperation to complete its form and meaning?  & 5 & 0 & 5 & 0 & 4 & 1.4 & 3.5 & 0.7 & 5 & 0 & 3.5 & 0.7 & 4.3 & 0.5 \\
				\hline
				7.-Does it propose a new vision of reality? & 2 & 1.4 & 2 & 1.4 & 1 & 0 & 2 & 1.4 & 2 & 1.4 & 2.5 & 0.7 & 1.9 & 1.1 \\
				\hline
				8.-Does it propose a new vision of the genre it uses? & 2 & 1.4 & 1.5 & 0.7 & 1 & 0 & 1 & 0 & 2 & 1.4 & 2 & 0 & 1.6 & 0.6 \\  
				\hline
				9.-Does it propose a new vision of the language itself? & 2 & 1.4 & 1 & 0 & 1 & 0 & 1 & 0 & 1 & 0 & 1 & 0 & 1.2 & 1.2\\  
				\hline
				\multicolumn{15}{c}{\textbf{Editorial / commercial}}  \\
				\hline
				10.-Does it remind you of another text or book you have read? & 4 & 1.4 & 2 & 1.4 & 1 & 0 & 5 & 0 & 4 & 1.4 & 4.5 & 0.7 & 3.4 & 0.8\\  
				\hline
				11.-Would you like to read more texts like this?  & 3 & 1.4 & 2 & 1.4 & 1 & 0 & 4 & 1.4 & 3 & 2.8 & 4 & 1.4 & 2.8 & 1.6 \\ 
				\hline 
				12.-Would you recommend it? & 3 & 2.8 & 1 & 0 & 1 & 0 & 4 & 1.4 & 3 & 2.8 & 4 & 1.4 & 2.7 & 1.4 \\
				\hline
				13.-Would you give it as a present? & 1 & 0 & 1 & 0 & 1 & 0 & 4 & 1.4 & 3 & 2.8 & 4 & 1.4 & 2.3 & 0.9 \\  
				\hline
			\end{tabular}
		\end{table}
		
		\begin{table}
			\scriptsize
			\caption{Literary Experts’ Evaluations of Microfictions Generated by Monterroso and ChatGPT-3.5 by GrAImes Section.}
			\label{table:lit_exp_mont_chatgpt_summ_psect}
			\begin{tabular}{c c c c} 
				\hline
				\multicolumn{4}{c} Story overview and text complexity  \\
				\hline
				\#& MF & AV & SD\\  
				\hline
				1& 4 & 4 & 1\\
				\hline
				2& 5 & 4 & 1\\
				\hline
				3& 6 & 3.5 & 2.1\\
				\hline
				4& 1 & 3 & 2.8\\
				\hline
				5& 2 & 3 & 2.8\\
				\hline
				6& 3 & 1 & 0\\
				\hline
			\end{tabular} \hspace{.2 cm}	
			\begin{tabular}{ c c c}
				\hline
				\multicolumn{3}{c}{\textbf{Technical}}  \\ 
				\hline
				MF & AV & SD\\ 
				\hline
				1 &3.2 & 0.8\\
				\hline
				5 & 3 & 0.6\\
				\hline
				6 & 2.8 & 0.3\\
				\hline
				4 & 2.5 & 0.4\\
				\hline
				2 & 2.5 & 1\\
				\hline
				3 & 1.8 & 0.6\\
				\hline
			\end{tabular} \hspace{.2 cm}
			\begin{tabular}{c c c} 
				\hline
				\multicolumn{3}{c}{\textbf{Editorial/commercial}}  \\
				\hline
				MF & AV & SD\\ 
				\hline
				4 & 4.3 & 1.1\\
				\hline
				6 & 4.1& 1.2\\
				\hline
				5 & 3.3 & 2.5\\
				\hline
				1 & 2.8 & 1.8\\
				\hline
				2 & 1.5 & 0.7\\
				\hline
				3 & 1 & 0\\
				\hline
			\end{tabular}\hspace{.2 cm}	
			\begin{tabular}{c c c} 
				\hline
				\multicolumn{3}{c}{\textbf{Microfictions Total Analysis}}  \\
				\hline				
				MF & AV & SD\\ 
				\hline
				4 & 3.4 & 0.8\\
				\hline
				6& 3.4 & 0.8\\
				\hline
				5 & 3.2 & 1.4\\
				\hline
				1 & 3 & 1.4\\
				\hline
				2 & 2.2 &1.1\\
				\hline
				3 & 1.4 & 0.3\\
				\hline
			\end{tabular}
		\end{table}
		
		\begin{figure}[]
			\caption{Line charts of literary experts GrAImes sections summarized AV and SD}
			\label{fig:line_chart_lit_exp_graimes_sections}
			\vspace{.2 cm}
			\includegraphics[width=7cm]{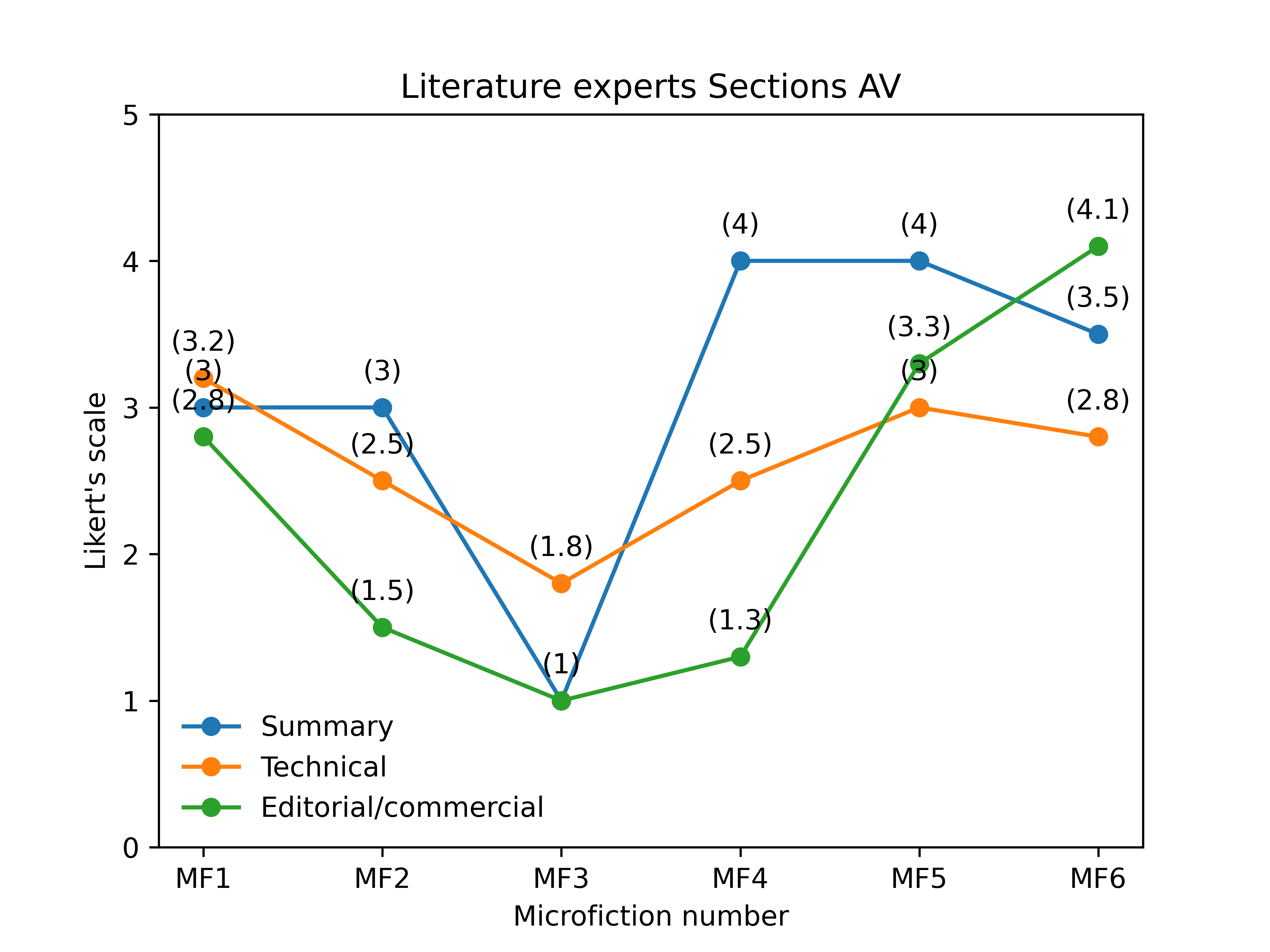}
			\includegraphics[width=7cm]{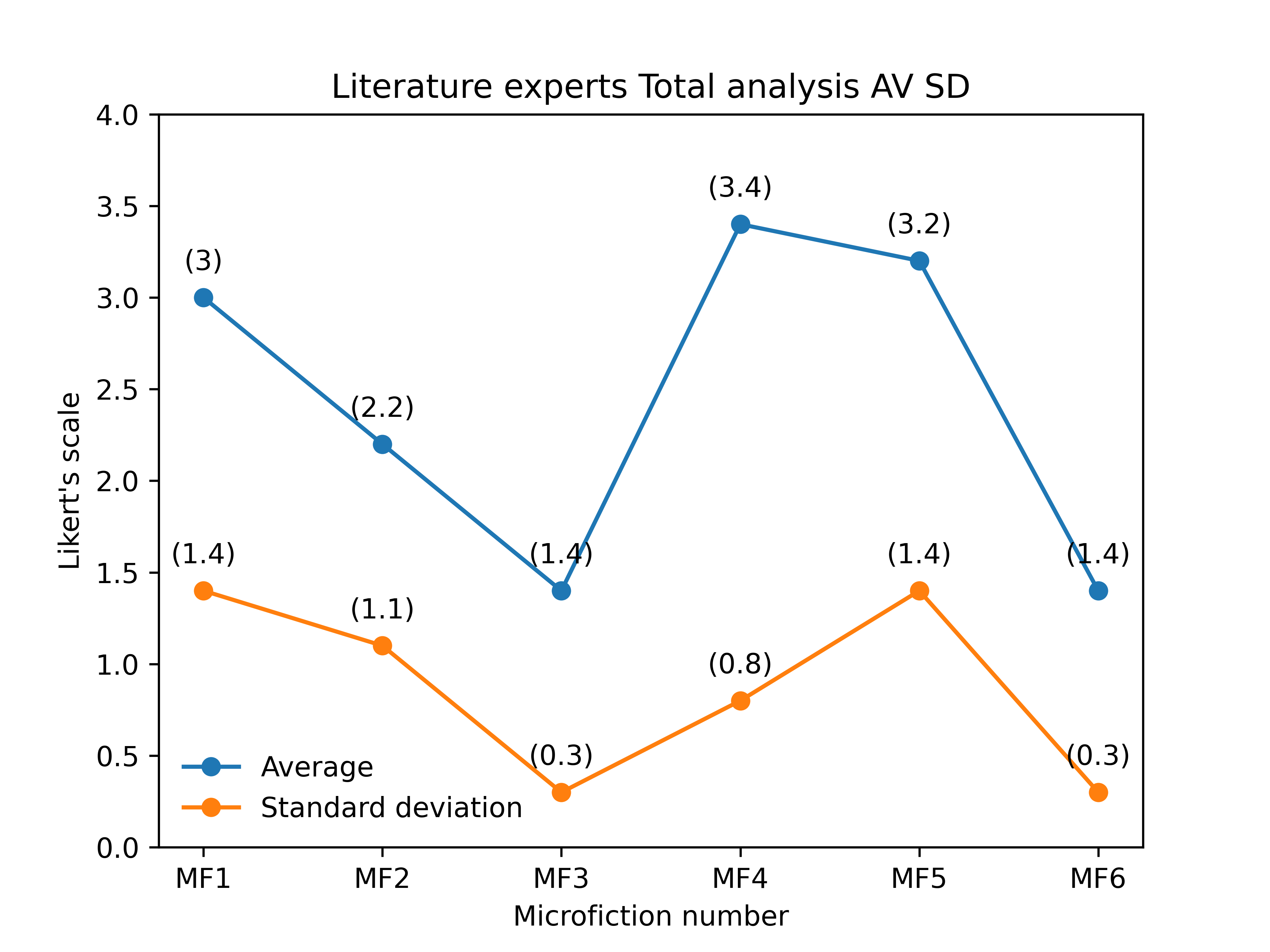}
		\end{figure}
				
		One of the most notable results in this evaluation concerns the interpretative engagement of readers. The highest-rated question, \textit{``Does the text require your participation or cooperation to complete its form and meaning?''}, received an average score (\textbf{AV}) of 4.3 with a standard deviation (\textbf{SD}) of 0.5. This suggests that the evaluated texts demand significant reader interaction, a crucial trait of literary complexity (see table \ref{table:litexps_quest_SD_order_mont_chatgpt}).
		
		Conversely, aspects related to innovation in language and genre were rated lower. The question \textit{``Does it propose a new vision of the language itself?''} obtained an AV of 1.2, the lowest among all items, with an SD of 1.2, indicating high variability in responses. Similarly, the question \textit{``Does it propose a new vision of the genre it uses?''} received an AV of 1.6 and an SD of 0.6, further emphasizing the experts' perception that the generated texts do not significantly redefine literary conventions.
		
		Regarding textual credibility, the question \textit{``Is the story credible?''} was rated highly, with an AV of 4.2 and an SD of 0.7. This suggests that the narratives effectively maintain verisimilitude, an essential criterion for reader immersion. Additionally, evaluators were asked whether the texts reminded them of other literary works, yielding an AV of 3.4 and an SD of 0.8, indicating a moderate level of intertextuality.
		
		GrAImes also examined subjective aspects of reader appreciation. The questions \textit{``Would you recommend it?''} and \textit{``Would you like to read more texts like this?''} received AV scores of 2.7 and 2.8, respectively, with higher SD values (1.4 and 1.6), reflecting diverse opinions among experts. Similarly, the willingness to offer the text as a gift scored an AV of 2.3 with an SD of 0.9, suggesting a moderate level of appreciation but not a strong endorsement.

		\begin{figure}[h!]
			\caption{Comparison of literary experts vs enthusiasts. GrAImes Sections AV with AI generated MFs.}			\label{fig:comp_exp_enth_mf_AI}
			\vspace{.2 cm}
	\includegraphics[width=.5\linewidth]{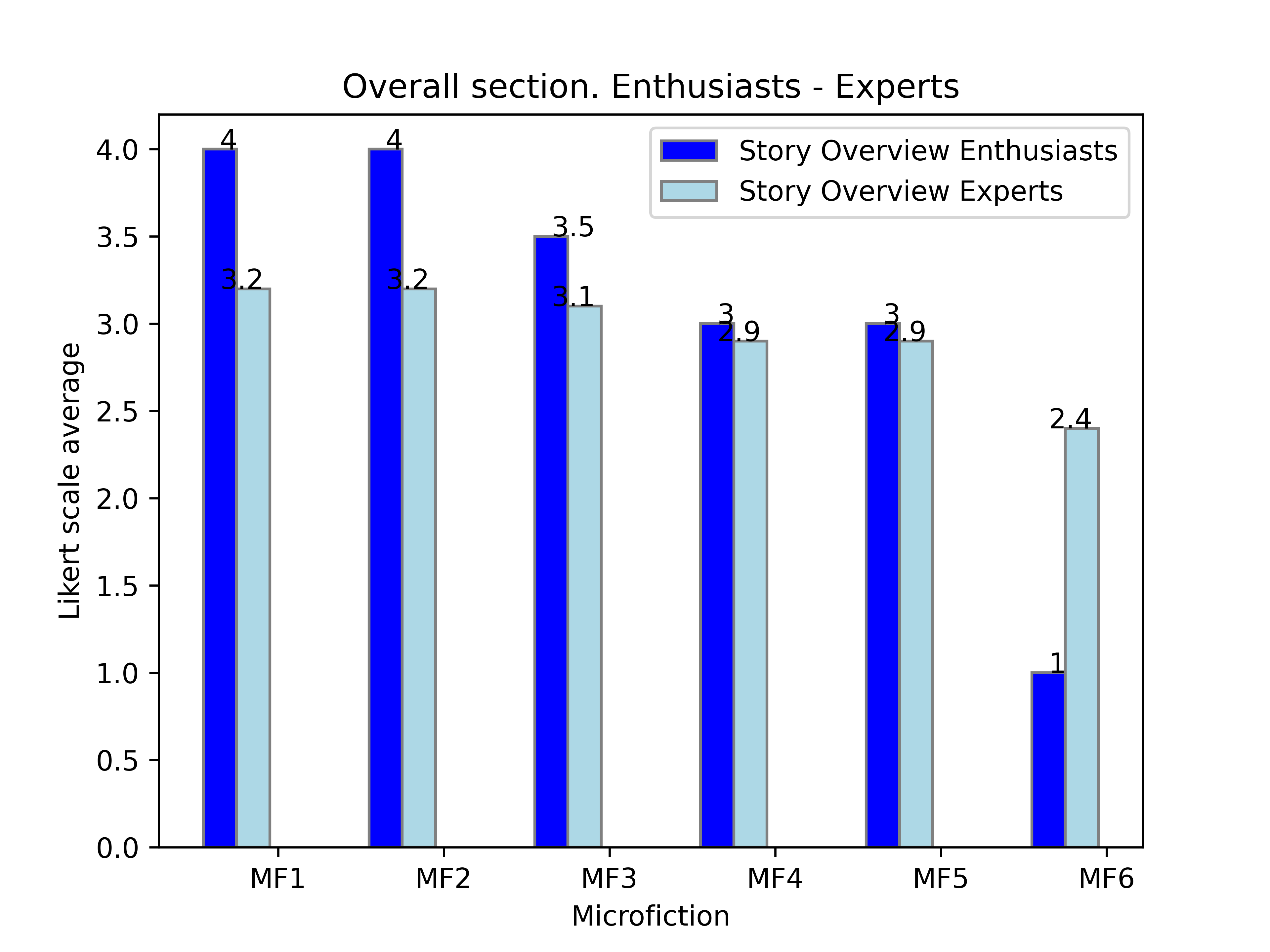}
        \includegraphics[width=.5\linewidth]{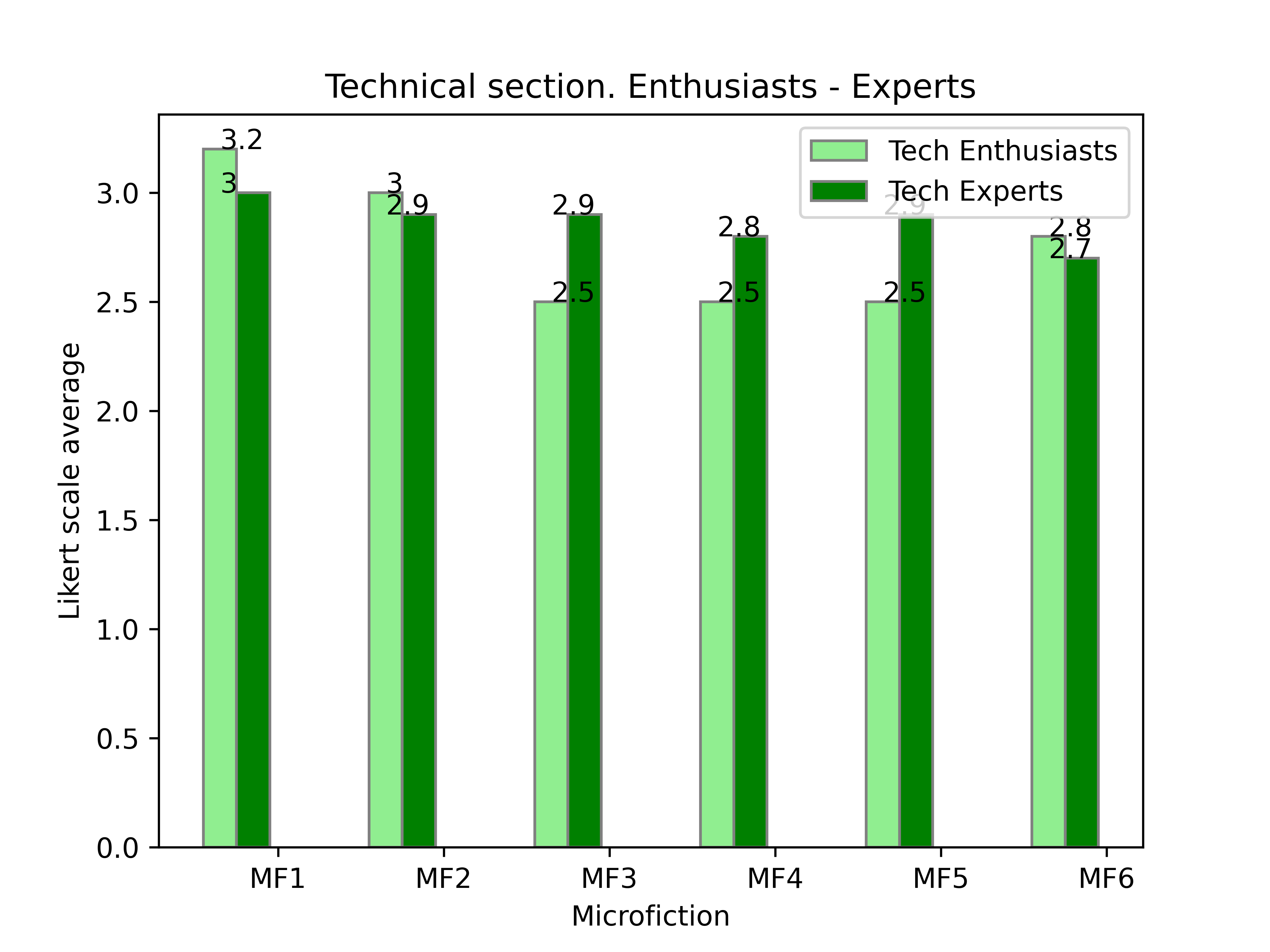}
        \includegraphics[width=.5\linewidth]{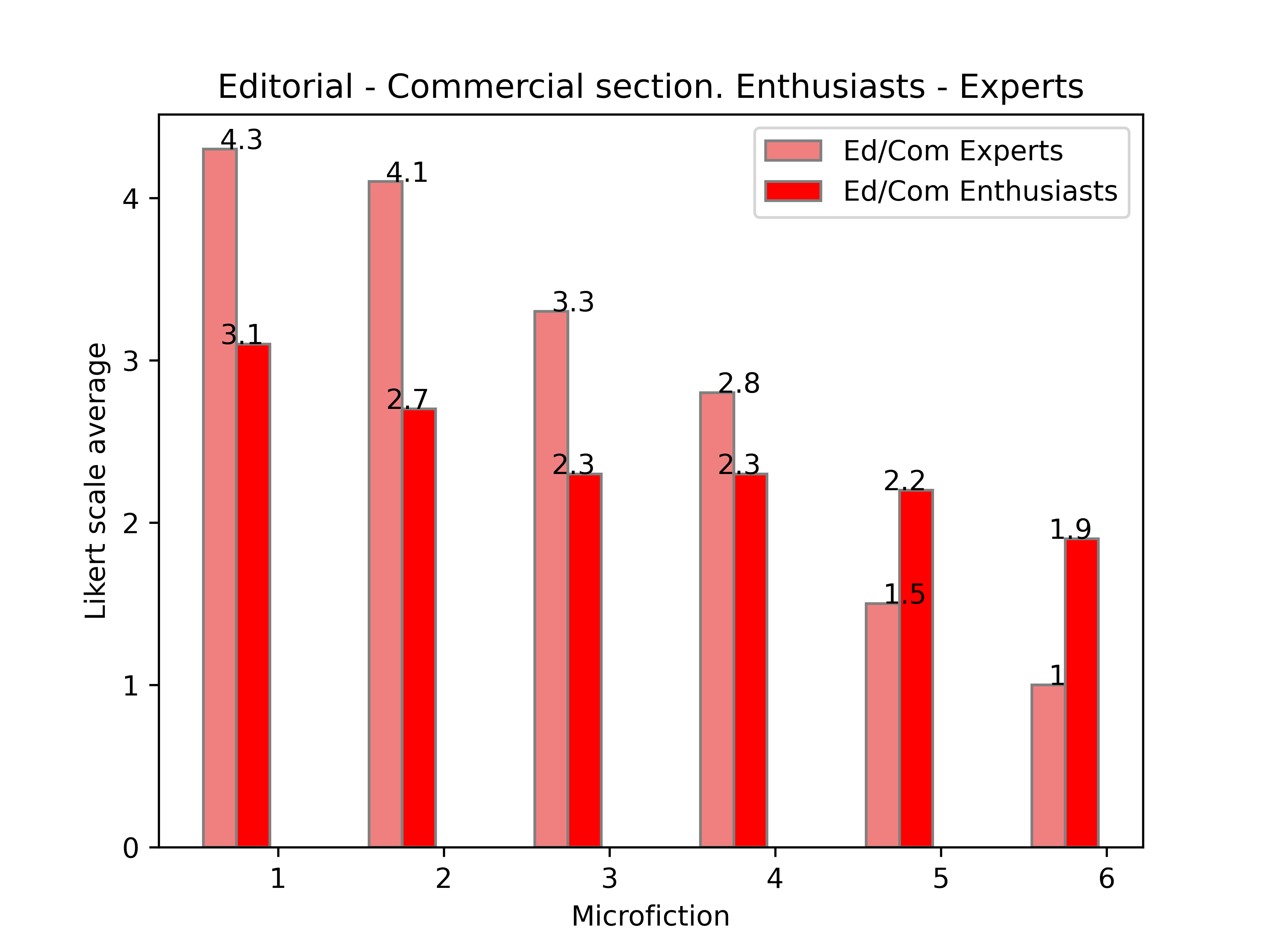}
        \includegraphics[width=.5\linewidth]{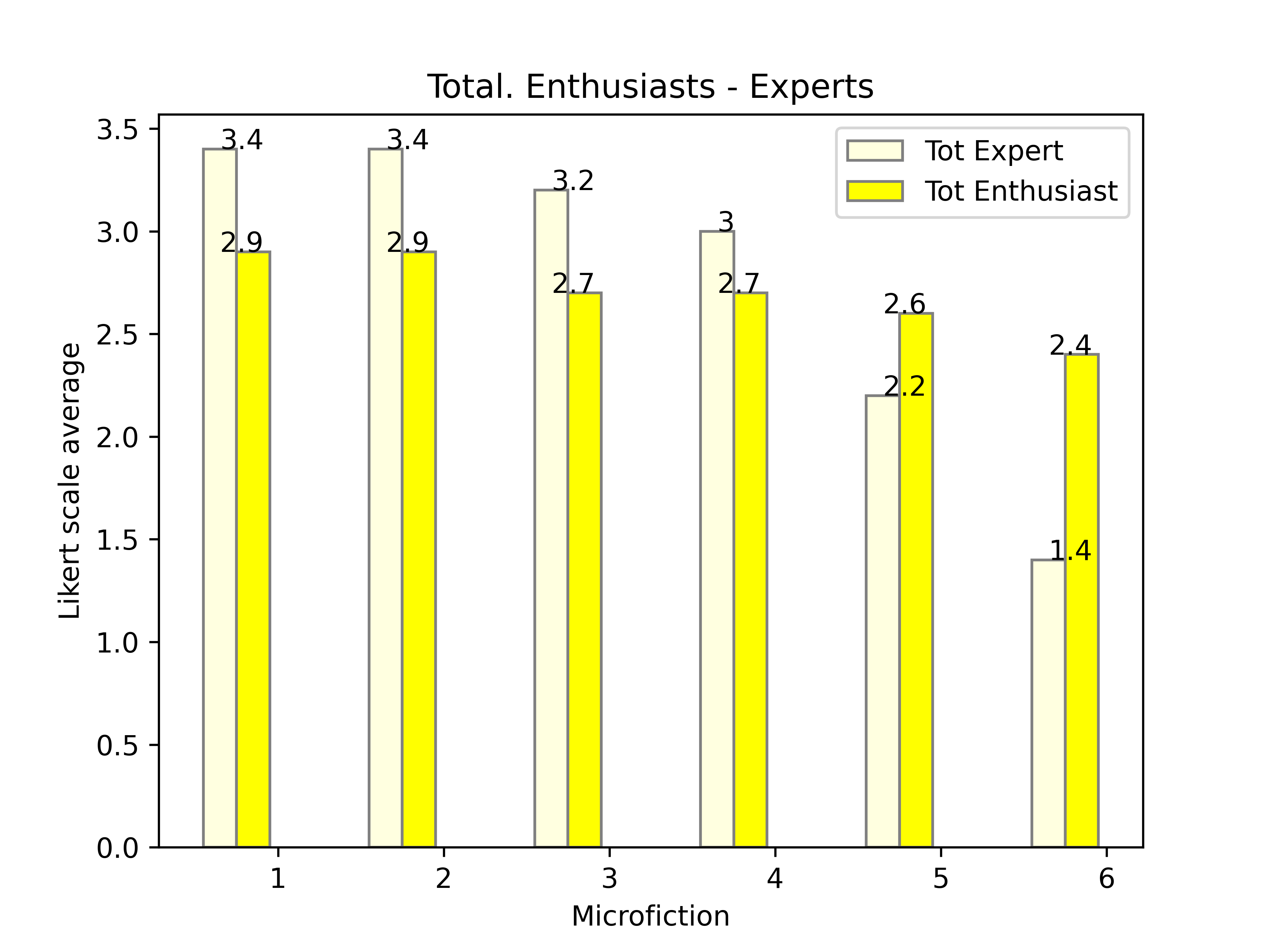}
		\end{figure}

		\begin{table}[h]
			\scriptsize
			\caption{Literary Experts’ Responses to Microfictions Generated by Monterroso and ChatGPT-3.5 Organized by Ascending Standard Deviation}
			\label{table:litexps_quest_SD_order_mont_chatgpt}	
			\textbf{}	\begin{tabular}{
					m{40em} c m{1em}
				} 
				\hline
				\multicolumn{3}{c}{\textbf{Literary experts responses to MF's generated by Monterroso and ChatGPT-3.5, ordered by SD}}\\
				\hline
				Question 
				& AV & SD \\
				\hline
				6.-Does the text require your participation or cooperation to complete its form and meaning? 
				&4.3 &0.5\\ 
				\hline
				8.-Does it propose a new vision of the genre it uses? 
				& 1.6 & 0.6\\
				\hline
				5.-Is the story credible?
				& 4.2 &0.7\\
				\hline
				10.-Does it remind you of another text or book you have read? & 
				3.4 &0.8\\ 
				\hline
				13.-Would you give it as a present? 
				&2.3 &0.9\\
				\hline			
				7.-Does it propose a new vision of reality? 
				& 1.9 &1.1\\
				\hline
				9.-Does it propose a new vision of the language itself? 
				& 1.2 &1.2\\
				\hline
				12.-Would you recommend it? 
				& 2.7 &1.4\\
				\hline
				11.-Would you like to read more texts like this?  
				& 2.8 &1.6\\
				\hline
				3.-Does it propose other interpretations, in addition to the literal one? 
				& 3.1 & 1.8 \\
				\hline  	
			\end{tabular}
		\end{table}

         There are currently no existing state-of-the-art references available for direct comparison with the present study, highlighting the novelty of our approach. This study introduces an innovative evaluation protocol for microfictions, validated by literary experts, and contributes to the field by assessing both human-written and AI-generated texts, specifically those produced by chatbots such as ChatGPT. The absence of direct SOTA comparisons is due to the lack of prior work that simultaneously addresses the evaluation of microfiction across these two distinct origins—human authors and generative AI. This gap in the literature underscores the significance of our research, which seeks to establish a comprehensive framework for assessing literary quality that is applies to both human and AI-generated works. The focus on this specific research area is justified by the increasing prominence of AI in literary production and the need for reliable, expert-validated evaluation tools to assess the literary merit of AI-generated texts in comparison to traditional human-authored literature.
        
		\section{Discussion}
			
		In our study, the evaluators were literary experts and literature enthusiasts, the evaluation of AI-generated and human written literary texts differs significantly depending on the expertise and evaluators reading habits. Literature scholars and dedicated literature enthusiasts possess a familiarity with narrative structures, stylistic devices, and literary traditions, enabling them to assess texts with a critical and informed perspective. They are more likely to recognize intertextual references, thematic depth, and the subtleties of language that contribute to literary quality. In contrast, evaluations conducted by enthusiast readers —most of whom engage with literature only occasionally— tend to focus on immediate readability, entertainment value, and emotional impact rather than on formal or aesthetic complexity. While this broader audience provides valuable insights into general reception and accessibility, their assessments may lack the depth needed to critically engage with intricate literary techniques. This divergence underscores the importance of distinguishing between different evaluator groups when developing assessment methodologies for AI-generated texts. A balanced evaluation framework should account for both perspectives, ensuring that AI literature is judged not only on its mass appeal but also on its adherence to, or innovation within, established literary traditions.
								
		The evaluation of AI-generated literary texts poses significant challenges due to the inherently subjective nature of aesthetic judgment. Traditional assessment frameworks in computational linguistics often rely on automated metrics such as perplexity, coherence, and sentiment analysis. However, these metrics fail to capture the nuanced and context-dependent qualities that define literary excellence. As a result, reader-based evaluation has emerged as a crucial methodological approach, leveraging human perception to assess the artistic and stylistic value of AI-generated narratives. The variability in responses between random readers and literary experts highlights the necessity of a structured framework that integrates both perspectives to ensure a more comprehensive and reliable evaluation process.
		
		A key aspect of aesthetic evaluation is the distinction between general audience reception and expert literary critique. While non-expert readers provide insights into accessibility, engagement, and emotional resonance, experts apply specialized knowledge of literary traditions, narrative structures, and stylistic innovation. This distinction becomes particularly relevant in the assessment of AI-generated texts, where algorithmic authorship often lacks intentionality and depth in meaning construction. Consequently, the presence of literary experts in the evaluation design process is not only beneficial but essential for identifying higher-order textual attributes such as intertextuality, originality, and thematic complexity.
		
		Given these considerations, a hybrid evaluation model of GrAImes that integrates both expert critique and broader audience participation is likely to yield the most balanced assessment of AI-generated literature. The inclusion of expert evaluators ensures that the texts are measured against established literary standards, while the involvement of general readers provides valuable feedback on accessibility and reader engagement. This dual approach underscores the need for interdisciplinary collaboration between computational linguists and literary scholars in the design of methodologies for AI and human written text evaluation. Ultimately, a hybrid evaluation model that harmonizes expert insight with audience feedback not only enriches the assessment of AI-generated literature but also fosters a collaborative dialogue between disciplines, paving the way for a more inclusive understanding of literary value.
		
		Expert consensus on the protocol's capacity to evaluate microfiction's literary value was predominantly positive, with 80\% of reviewers affirming its effectiveness. However, the presence of a dissenting perspective underscores the importance of continuous methodological refinement. The minor reservations primarily centered on the precision of specific evaluation criteria, indicating a nuanced approach to protocol development is required. In summary, while the strong endorsement from experts highlights the protocol's promise, the dissenting voices serve as a crucial reminder that ongoing refinement is essential to ensure that the evaluation of microfiction remains both precise and relevant.

		The evaluation protocol encountered significant methodological challenges, particularly regarding criterion ambiguity. Key issues included interpretative inconsistencies in assessing linguistic creativity, intertextual references, and expressive quality. Experts specifically highlighted problematic areas such as the subjective interpretation of “novel language“ and the complex evaluation of literary influences. The recommended methodological improvements include replacing vague, recall-based assessments with more explicit, structured inquiries about literary lineage and explicit criteria for measuring expressive quality.In further experiments we aim to tackle these methodological challenges by implementing clearer and more structured criteria, as this will significantly enhance the evaluation protocol and ensure that the assessment of literary creativity and influence is conducted with both rigor and significance.

		Both groups found ChatGPT-3.5’s MFs more commercially appealing, while Monterroso’s texts showed slightly better technical execution (literature enthusiasts) or reader engagement (experts). These findings highlight that while AI-generated microfictions can compete with human-authored ones in commercial and structural aspects, they still fall short in innovative and deeply interpretive literary qualities. The study underscores the potential of AI in replicating certain narrative techniques while emphasizing the enduring challenges of achieving true creative originality. Experts focused on depth and originality, while literature enthusiasts prioritized readability and appeal, reinforcing that AI-generated texts may satisfy casual readers more than experts in literature. In essence, while AI-generated microfictions demonstrate commercial viability and structural competence, they ultimately struggle to match the depth and originality that defines truly exceptional literature, revealing a persistent gap between algorithmic creation and human artistry.

               Deep lexical and semantic comparison metrics, such as BERT-based measures, have been employed to approximate human judgment in poetry generation by assessing lexical diversity and semantic similarity \citep{lo2022gpoet}. However, these metrics were originally developed for tasks like automatic translation and summarization (BLEU, ROUGE), where a reference "gold standard" text (e.g., a human translation or summary) exists. Given the highly subjective nature of evaluating literary qualities, such as depth, originality, innovation, disruption, and structural complexity, and the lack of gold standard text, BERT based measures alone may not be a sufficient measure for assessing these dimensions.

                Incorporating literary theory and engaging readers with expertise in literature (whether scholars or enthusiasts) could challenge recent findings suggesting that AI-generated poetry is indistinguishable from human-written classical poetry \citep{porter2024ai}. This is particularly relevant when evaluators are sourced from anonymous crowdsourcing platforms like Prolific \footnote{\url{https://www.prolific.com/}}. If reader experience significantly influences text reception, the results of poetry evaluation experiments might differ substantially when conducted with literary experts or enthusiasts rather than general participants.
               
                Our results indicate that integrating literary knowledge and evaluators familiar with literature (or even experts) into the evaluation protocol enhances the assessment of texts produced by experienced human authors. Additionally, this approach improves the evaluation of microfictions generated by more advanced, well-trained generative language models.

		\section{Conclusions}
		
        This study introduces a literary evaluation protocol for human-written and AI-generated microfictions, integrating literary theory and expert input to ensure rigorous assessment of narrative quality, stylistic coherence, and creative depth. By grounding the framework in established literary principles, it enhances adaptability across genres while improving validity over superficial metrics. The inclusion of domain-specific expertise enables meaningful comparisons between human and AI-generated texts, offering a scalable tool for computational creativity research.
        	        

        GrAImes revealed divergent evaluative priorities: literary experts emphasized technical execution and originality, while enthusiasts favored accessibility and enjoyment. ChatGPT-3.5’s high-data training yielded coherent but unoriginal outputs, whereas Monterroso’s limited dataset produced inconsistent results. Standard deviation analysis showed strong expert consensus (SD $\approx$ 0) versus moderate enthusiast variability (SD $\approx$ 1.3–1.4), highlighting how evaluator background shapes perception of literary quality.

        GrAImes systematically benchmarked AI systems, exposing gaps like innovation deficits and training-data dependencies. While future refinements could integrate qualitative open-ended questions, the current framework successfully measures AI microfiction performance, demonstrating that evaluation design critically influences interpretations of computational creativity.
        
        A forthcoming experiment will apply the protocol to chatbot-generated microfictions, incorporating expert, reader, and chatbot self-assessments. Experts will evaluate literary nuance, readers will assess engagement, and AI self-critiques will provide comparative insights. This multi-perspective approach aims to validate the protocol’s robustness while exploring discrepancies between human and machine evaluative standards, advancing computational literary analysis.
	
		
		
		\vspace{6pt} 
		
		
		
		
		
		\authorcontributions{
			Conceptualization, J.G.F., R.M., G.A.M., N.P., N.C.A., and Y.G.M.; methodology, R.M. and J.G.F.; software, G.A.M.; validation, G.A.M., R.M. and J.G.F.; formal analysis, R.M. and J.G.F.; investigation, G.A.M. and J.G.F.; resources, R.M. and J.G.F.; data curation, G.A.M. and J.G.F.; writing---original draft preparation, G.A.M.; writing---review and editing, R.M., J.G.F., G.A.M., N.P., N.C.A. and Y.G.M.; visualization, G.A.M.; supervision, R.M. and J.G.F.; project administration, R.M.; funding acquisition, R.M. All authors have read and agreed to the published version of the manuscript.
			}
			
			\funding{This research was funded by ECOS NORD grant number 321105.}
			
			
			\informedconsent{Informed consent was obtained from all subjects involved in the study.}
				
			
			\dataavailability{All materials needed to replicate the experiment are available at: \url{https://github.com/Manzanarez/GrAImes}.}

			\acknowledgments{
				 We are grateful to the writers, literary experts and enthusiast readers who participated in this research: Miguel Tapia, Florance Olivier, Oswaldo Zavala, Marcos Eymar, Alejandro Lambarry, Dennis G. Wilson, Abraham Truxillo, Abril Albarran, Adriana Azucena Rodriguez, David Nava, Lupita Mejia Alvarez, Sandra Huerta, Maria Elisa Leyva Gonzalez, Maria Mendoza, Luis Roberto, Diana Leticia Portillo Rodriguez, Angelica, Iris, Elisa, Fernanda, Guadalupe Monserrat Ramirez Santin, Valeria, Janik Rojas, Brenda, Alma Sanchez and Jorge Luis Borges}\footnote{Jorge Luis Borges (1899–1986) was an Argentine writer, poet, and essayist widely regarded as one of the most influential literary figures of the 20th century. Known for his intricate short stories exploring themes such as infinity, mirrors, labyrinths, and the nature of authorship, Borges played a foundational role in modern literature and philosophical fiction. Among Spanish-speaking readers, he is often considered, alongside Miguel de Cervantes, one of the two most important authors in the history of the Spanish language.}.
			
			\conflictsofinterest{The authors declare no conflicts of interest.
				
				
				
					
				}
				
				


					\begin{adjustwidth}{-\extralength}{0cm}
					
					\reftitle{References}
					
					
					\bibliography{Definitions/references}
					\isChicagoStyle{%
						
					}{}
					
					\isAPAStyle{%
						
					}{}
					
					%
					
					
					\PublishersNote{}
					\end{adjustwidth}
			\end{document}